\documentclass{tdp}

\usepackage{graphicx}%
\usepackage{multirow}%
\usepackage{amsmath,amssymb,amsfonts}%
\usepackage{amsthm}%
\usepackage{mathrsfs}%
\usepackage[title]{appendix}%
\usepackage{xcolor}%
\usepackage{textcomp}%
\usepackage{manyfoot}%
\usepackage{booktabs}%
\usepackage{algorithm}%
\usepackage{algorithmicx}%
\usepackage{algpseudocode}%
\usepackage{listings}%

\usepackage{hyperref}
\usepackage{natbib}
\setcitestyle{aysep={}} 
\usepackage[caption=false]{subfig}
\usepackage{pdfpages}
\usepackage{mathtools}
\usepackage{rotating}

\begin{document}

\title{Bayesian Generative Adversarial Networks via Gaussian Approximation for Tabular Data Synthesis}
\author{Bahrul Ilmi Nasution$^{1*}$, Mark Elliot$^{1}$, Richard Allmendinger$^{2}$}
\address{$^{1}$Department of Social Statistics, The University of Manchester, Oxford Rd, Manchester, M13 9PL, UK. \\ 
  $^{2}$Alliance Manchester Business School, The University of Manchester, Booth Street West, Manchester, M15 6PB, UK.\\
  $^{*}$ Corresponding Author \\
  E-mail: {\small \tt{firstname.lastname@manchester.ac.uk}}
}

\TDPRunningAuthors{Bahrul Ilmi Nasution, Mark Elliot, Richard Allmendinger}
\TDPRunningTitle{Gaussian-Approx GAN for Tabular Data}
\TDPThisYear{2026}
\TDPFirstPageNumber{1}
\TDPSubmissionDates{Received 17 July 2025; received in revised form 12 January 2026; accepted 24 February 2026}

\maketitle

\begin{abstract}
Generative Adversarial Networks (GAN) have been used in many studies to synthesise mixed tabular data. Conditional tabular GAN (CTGAN) have been the most popular variant but struggle to effectively navigate the risk-utility trade-off. Bayesian GAN have received less attention for tabular data, but have been explored with unstructured data such as images and text. The most used technique employed in Bayesian GAN is Markov Chain Monte Carlo (MCMC), but it is computationally intensive, particularly in terms of weight storage. In this paper, we introduce Gaussian Approximation of CTGAN (GACTGAN), an integration of the Bayesian posterior approximation technique using Stochastic Weight Averaging-Gaussian (SWAG) within the CTGAN generator to synthesise tabular data, reducing computational overhead after the training phase. We demonstrate that GACTGAN yields better synthetic data compared to CTGAN, achieving better preservation of tabular structure and inferential statistics with less privacy risk. These results highlight GACTGAN as a simpler, effective implementation of Bayesian tabular synthesis.
\end{abstract}


\section{Introduction}
\label{sec:intro}

Synthetic data has become an increasingly valuable asset across multiple domains, offering practical solutions to privacy-preserving data sharing. It facilitates the release of data by mitigating disclosure risks~\citep{Raab2024privacy}, and supports pedagogical use cases by enabling students to work with realistic yet non-disclosive datasets~\citep{Elliot2024production,little2025producing}. However, high-dimensional tabular data-such as that from censuses and social surveys-poses significant challenges due to the presence of sensitive attributes (e.g., demographic, health, financial). As a result, data controllers often impose strict access restrictions, limiting broader reuse, including in educational settings. To address this, researchers—particularly within national statistical offices (NSOs)—have developed synthetic data generation methods aimed at preserving key statistical properties while reducing disclosure risk~\citep{nowok2016synthpop1,Little2023federated}.

Synthetic data generation is commonly pursued through two broad methodological paradigms: statistical and deep learning-based approaches. Statistical methods assume that the underlying data distribution can be effectively captured by a single model, with examples including Classification and Regression Trees (CART)~\citep{nowok2016synthpop1} and Bayesian Networks (BN)~\citep{Zhang2017privbayes}. In contrast, deep learning approaches—often referred to collectively as deep generative models (DGMs)—leverage neural networks to approximate complex, high-dimensional data distributions. Building on the universal approximation theorem, which states that sufficiently deep neural networks can approximate any continuous function~\citep{Hornik1989multilayer}, DGMs have become a vibrant area of research~\citep{goodfellow2014generative,ma2020vaem,xu2019modeling}.

Among these, generative adversarial networks (GAN)~\citep{goodfellow2014generative} are among the most prominent, producing high-quality samples via adversarial training between generator and discriminator networks. However, standard GAN are susceptible to \textit{mode collapse}, where the generator fails to capture the full diversity of the data distribution~\citep{saatci2017bayesian}. This challenge is particularly acute for tabular data, which often comprises heterogeneous feature types (e.g. numerical and categorical variables), introducing complex optimisation landscapes~\citep{ma2020vaem,nasution2022data}. Conditional tabular GAN (CTGAN)~\citep{xu2019modeling} addresses some of these challenges by modelling conditional distributions and using training-by-sampling techniques, making it a promising candidate for synthesising structured data. Nonetheless, concerns persist around the trade-off between utility and disclosure risk in synthetic outputs~\citep{ran2024multiobjective}.

Bayesian neural networks (BNN) offer a potentially compelling alternative by estimating posterior distributions over model parameters, rather than relying on fixed point estimates as in classical neural networks. In the context of DGMs, this Bayesian treatment enables exploration of multiple modes in the data distribution, potentially improving sample diversity~\citep{saatci2017bayesian}. Bayesian DGMs typically approximate the intractable posterior via methods such as Markov Chain Monte Carlo (MCMC)~\citep{saatci2017bayesian,Gong2019icebreaker,turner2018metropolis,salimans2015markov} or variational approximation~\citep{glazunov2022do,tran2017hierarchical}. While MCMC approaches provide strong asymptotic guarantees, they are computationally intensive and memory-demanding, as they require storing multiple samples of the model weights across iterations.

A more tractable approach to posterior approximation involves projecting the network weights into a multivariate normal distribution defined by a mean vector and covariance matrix. This naturally raises the question: how should the mean and covariance be specified? The Laplace approximation~\citep{MacKay1992a}, which estimates the covariance using second-order derivatives of the loss function, is a well-established method. However, it is computationally expensive, with a cost that scales quadratically with the number of parameters~\citep{daxberger2021laplace}. Simplified diagonal approximations—such as those based on the inverse Fisher information—are more efficient but can produce unstable estimates, including exploding variances, especially in settings where vanishing gradients are common, such as GAN training~\citep{gulrajani2017improved}.

To address these limitations, stochastic weight averaging–Gaussian (SWAG)~\citep{maddox2019simple} has been proposed as a scalable and efficient alternative. SWAG approximates the posterior by capturing the trajectory of weights over the course of training to estimate both the mean and covariance, offering a lightweight yet effective method for posterior inference. When applied to GAN, SWAG enables a Bayesian treatment of the generator with minimal additional computational burden compared to traditional methods such as gradient averaging or second-order techniques~\citep{Murad2021probabilistic}.

While GAN, including their Bayesian variants~\citep{saatci2017bayesian, mbacke2023pacbayesian}, have been used primarily in image and text applications, the implementation for tabular data synthesis, particularly using SWAG, remains largely unexplored. Existing studies have focused on applying weight averaging techniques in GAN for tasks such as image segmentation~\citep{Durall2019object} but have not fully integrated SWAG into GAN frameworks for tabular data. However, SWAG has been successfully applied in a variety of other real-world domains, such as field temperature prediction~\citep{Morimoto2022assessments}, air quality monitoring~\citep{Murad2021probabilistic}, and earthquake fault detection~\citep{Mosser2022a}.

Therefore, this study helps connect Bayesian modeling with tabular data synthesis. Our contributions are threefold:

\begin{enumerate}
\item \textbf{GACTGAN: Bayesian posterior over the generator.} We introduce Gaussian Approximation of CTGAN (GACTGAN), which integrates SWAG with CTGAN to approximate a posterior distribution for the generator parameters. (Section~\ref{sec:swagctgan})

\item \textbf{Practical implementation.} We provide an implementable version of GACTGAN based on different configurations of SWAG. (Sections~\ref{ssec:swactgan-impl})

\item \textbf{Improved utility with lower risk.} We show that GACTGAN strengthens CTGAN, producing synthetic data that remains similarly useful while reducing risk relative to standard CTGAN. (Section~\ref{sec:results})
\end{enumerate}

We view this work as a step toward making Bayesian methods more actionable for tabular data synthesis, and we hope it will enable further research as well as practical deployments.

This paper is structured as follows. Section~\ref{sec:rw} provides background information on CTGAN and SWAG. Section~\ref{sec:swagctgan} formulates our proposed method, GACTGAN. Section~\ref{sec:method} describes the methodology, including the dataset, the evaluation, and the environmental setting. Section~\ref{sec:results} presents the results of our experiments. Finally, Sections~\ref{sec:concl} discuss limitations, future work, and conclusions.

\section{Background}
\label{sec:rw}

This section provides an initial overview before diving into our proposed methodology. GACTGAN is composed of two main elements: the CTGAN and the posterior approximation facilitated by SWAG. In Subsection~\ref{ssec:ctgan}, we give a concise introduction to CTGAN. Next, we explore the weight averaging methods in Subsection~\ref{ssec:wa}, followed by a posterior formulation using SWAG in Subsection~\ref{ssec:swag}.

\subsection{CTGAN}
\label{ssec:ctgan}

Generative Adversarial Networks (GAN) are DGMs that capture data distribution via an adversarial game between two neural networks~\citep{goodfellow2014generative}. The generator network, denoted as $G$ and parameterised by $\theta_G$, aims to generate realistic data from random noise $z$ to deceive its adversary. The discriminator, denoted as $D$ with parameters $\theta_D$, strives to distinguish real from fake data. Both networks engage in a minimax game to reach equilibrium, creating high-quality synthetic data. The original GAN developed by~\cite{goodfellow2014generative}, also known as vanilla GAN, uses binary entropy loss. However, vanilla GAN suffers from unstable training and mode collapse, where the generator produces similar outputs, leading to low-diversity data. \cite{arjovsky2017wasserstein1} and~\cite{gulrajani2017improved} introduced the Wasserstein loss to address these issues, enhancing model stability and reducing mode collapse. For batch data $x$ sampled from the original dataset $\mathcal{D}$, the discriminator and generator weights are updated using Equation~\eqref{eq:vgan} for vanilla GAN and~\eqref{eq:wgan} for Wasserstein GAN. 

\begin{equation}
\label{eq:vgan}
(\widehat{\theta}_G,\widehat{\theta}_D) = \arg \min_{\theta_G} \max_{\theta_D} \mathbb{E}_x [\log D(x)]+\mathbb{E}_z [\log (1-D(G(z))]
\end{equation}

\begin{equation}
\label{eq:wgan}
    (\widehat{\theta}_G,\widehat{\theta}_D) = \arg \min_{\theta_G} \max_{\theta_D} \mathbb{E}_x [D(x)] - \mathbb{E}_z [D(G(z))]
\end{equation}

~\cite{xu2019modeling} developed conditional tabular GAN (CTGAN) for mixed data types in tabular data. CTGAN tackles challenges of mixed data types using a conditional generator that synthesises data based on categorical column values, enabling the model to learn variable dependencies. CTGAN efficiently synthesises tabular data, applied in fields for data simulation and preservation of privacy~\citep{Athey2024using,elliot2023samples}.

\subsection{Stochastic Weight Averaging}
\label{ssec:wa}

The concept of averaging weights in neural networks has gained traction, offering a novel perspective for identifying optimal solutions, referred to as Stochastic Weight Averaging (SWA)~\citep{Izmailov2018AveragingWL}. In short, SWA leverages the historical trajectory of weights optimised using standard algorithms, such as Stochastic Gradient Descent (SGD) and Adam, combined with a scheduled learning rate.

Standard optimisation methods often converge to a ``sharp minimum''—a narrow solution where small changes in the data or weights can cause a significant drop in performance. SWA addresses this by computing the arithmetic mean of the weights accumulated along the optimisation path~\citep{hwang2021adversarial}. Intuitively, while standard SGD traverses the edges of a low-loss region, averaging these points identifies the centroid of that region. This allows the model to settle into a ``flat minimum''—a wider, more stable region in the loss landscape. Solutions found in flat minima tend to generalise better to unseen data because they are less sensitive to minor shifts in the underlying distribution~\citep{Izmailov2018AveragingWL}.

Typically, the SWA process starts after several iterations of standard training. For simplicity, the calculation of the average weights $\overline{w}$ can be made at all specific intervals $c$. The choice of $c$ offers flexibility and can be tailored according to the specific requirements of the user. Denoting $n_\text{mod}$ as the total number of models that have been averaged so far, the update in iteration $t$ can be performed when $t\pmod{c} = 0$ using Equation~\eqref{eq:wswa2}.

\begin{equation}
    \label{eq:wswa2}
    \overline{w} = \frac{\overline{w}\times n_\text{mod} + w^{(t)}}{n_\text{mod} + 1}
\end{equation}

Due to the straightforward implementation, there are many studies that have used SWA to improve the performance of their neural networks, such as in adversarial learning~\citep{Kim2023bridged} and medical imaging~\citep{Pham2021ear,Yang2023stochastic}.

\subsection{Being Bayesian: SWA and Gaussian approximation}
\label{ssec:swag}

SWA-Gaussian (SWAG) is an extension of SWA with Bayesian perspective that captures uncertainty in the weight trajectory~\cite{maddox2019simple}. SWAG approximates the posterior distribution of weights using a multivariate normal distribution $\mathcal{N}(\mu, \Sigma)$, where $\mu$ and $\Sigma$ denote the mean of SWA and the approximate covariance matrix, respectively~\citep{glazunov2022do}. The covariance matrix is composed of both diagonal and non-diagonal elements. The diagonal covariance $\Sigma_\text{diag}$ is calculated using the mean and second sample moment, denoted $\overline{w^2}$. The details of the calculation can be seen in Equations~\eqref{eq:diagcov} and~\eqref{eq:wsquare}~\citep{onal2024gaussianstochasticweightaveraging}.

\begin{equation}
    \label{eq:diagcov}
    \Sigma_\text{diag} = \text{diag}(\overline{w}^2 - \overline{w^2})
\end{equation}

\begin{equation}
    \label{eq:wsquare}
    \overline{w^2} = \frac{\overline{w^2}\times n_\text{mod} + (w^{(t)})^2}{n_\text{mod} + 1}
\end{equation}

The diagonal covariance matrix is frequent in the neural network, particularly in the variational approximation~\citep{blundell2015weight}. However, its use is often too restrictive, as it fails to consider the relationship among random variables that are characterised by the covariance matrix. An approach to obtain the covariance matrix involves utilising the outer products of the deviations, as illustrated in Equation~\eqref{eq:cov1}.

\begin{equation}
    \label{eq:cov1}
    \Sigma\approx\frac{1}{n_\text{mod}-1}\sum_{i=1}^{n_\text{mod}}(w_i-\overline{w})(w_i-\overline{w})^\intercal=\frac{1}{n_\text{mod}-1}\widehat{D}\widehat{D}^\intercal
\end{equation}

$\widehat{D}$ represents the deviation matrix structured in a column format. Unfortunately, storing a full rank $\Sigma$ is prohibitively expensive~\citep{glazunov2022do}. A viable alternative is to use the low-rank covariance $\Sigma_\text{l-rank}$, which efficiently captures only the last K-rank of the deviation matrix. The use of a low-rank approximation proves to be practically advantageous in enhancing the performance of the SWAG model~\citep{maddox2019simple}. The $\Sigma_\text{l-rank}$ can be calculated using Equation~\eqref{eq:covl}.

\begin{equation}
    \label{eq:covl}
    \Sigma_\text{l-rank}=\frac{1}{K-1}\widehat{D}\widehat{D}^\intercal
\end{equation}

The approximate posterior distribution of the model weights is represented by a Gaussian distribution with a mean from SWA and its approximate covariance, expressed as $\mathcal{N}(\overline{w}, \alpha_G(\Sigma_\text{diag} + \Sigma_\text{l-rank}))$. By default, the scaling factor for the covariance, denoted as $\alpha_G$, is set to 0.5. However, users have the flexibility to choose alternative scaling factors, with recommendations from~\citep{maddox2019simple} suggesting values $\alpha_G\leq 1$. Although implementation may appear straightforward within the Bayesian context, practical applications of SWAG remain niche, being particularly pertinent in specific areas such as out-of-distribution detection~\citep{glazunov2022do} and the development of large language models (LLMs)~\citep{onal2024gaussianstochasticweightaveraging}.

\section{GACTGAN}
\label{sec:swagctgan}

\subsection{Model Formulation}

\begin{figure}[ht]
    \centering
    \includegraphics[trim={0.2cm 0.2cm 0.2cm 0.2cm}, clip, width=\textwidth]{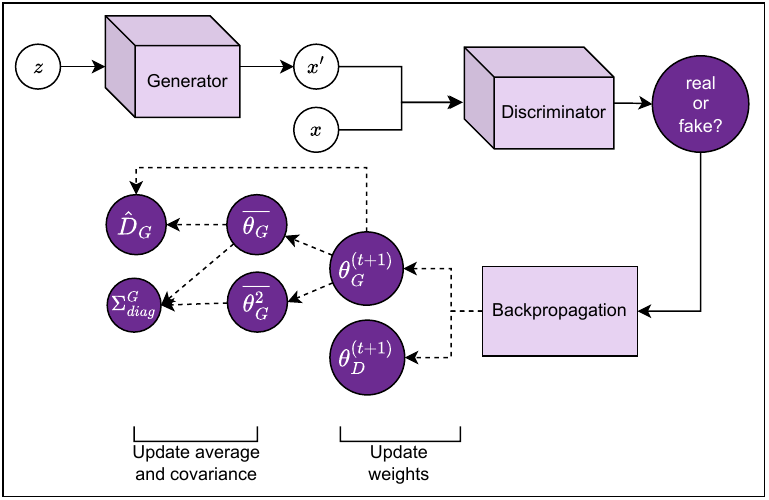}        
    \caption{An iteration process of GACTGAN. Note that the generator and discriminator is trained regularly. However, after the backpropagation in the generator, the weights are stored to update the mean and mean of squared weights. The mean and new weights are used to store the deviation matrix, while the diagonal covariance is constructed from the mean and mean of squared weights.}
    \label{fig:swagan}
\end{figure}

Figure~\ref{fig:swagan} provides an overview of an iteration of the training of the GACTGAN model, and Algorithm~\ref{alg:swagctgan} illustrates the details of the training process of the GACTGAN model in the form of pseudocode. In this study, we adopted the CTGAN methodology by~\cite{xu2019modeling}, with additional SWAG steps each epoch~\citep{maddox2019simple}. As illustrated in the figure, during the training phase, a noise $z$ is sampled from $\mathcal{N}(0,\textbf{I})$. Subsequently, this noise is fed into the $G$, resulting in the creation of synthetic data, denoted $x'$. The role of $D$ is to determine whether the data is real or fake by comparing them with the original dataset, which is selected according to the given batch size and labelled $x$. Consequently, Steps 7 of Algorithm~\ref{alg:swagctgan} update the discriminator and generator using gradients derived from the selected objective function: the vanilla loss (Equation~\eqref{eq:vgan}) or the Wasserstein loss (Equation~\eqref{eq:wgan}), as defined by the user. 

During the SWAG phase, the updated $\theta_G$ is stored to iteratively update the first and second samples every $c$. At the same time, the differences between the updated $\theta_G$ and the mean value $\overline{\theta}_G$ are used to estimate the deviation matrix. Given constraints on the low-rank approximation, if this maximum rank is surpassed, the oldest column is replaced by the newest addition. Calculating the covariance matrix using Equations~\eqref{eq:diagcov} and~\eqref{eq:covl} is carried out once training is complete, allowing the approximate posterior distribution of the generators to be derived using normal distribution.

\begin{algorithm*}[ht]
    \caption{GACTGAN}
    \label{alg:swagctgan}
    \begin{flushleft}
    \textbf{Input: } $\mathcal{D}$, $T$, epoch to start weight mean and covariance collection $t_\text{collect}$, $K$, Generator layer $G$, Discriminator layer $D$, batch size $M$ 
    \end{flushleft}
    \begin{algorithmic}[1]
    \State Obtain $x_\text{num}$ and $x_\text{dc}$ by transforming the numerical dataset using standard scaling based on Gaussian mode
    \State Obtain $x_\text{ohe}$ by transforming the categorical dataset using a hot encoding
    \State $x = x_\text{num} \oplus x_\text{dc} \oplus x_\text{ohe}$
    \State Sample data into $M$ number of mini-batches
    \For {$t=1\rightarrow T$}
        \For {$m=1\rightarrow M$}
            \State Update $\theta_D$ and $\theta_G$ using regular CTGAN procedure~\citep{xu2019modeling}
        \EndFor
        \If{$t > t_\text{collect}$}
            \State Update $\overline{\theta}_G$ using Equation~\eqref{eq:wswa2} and $\overline{\theta^2}_G$ using Equation~\eqref{eq:wsquare} by setting $w^{(t)}=\theta_G^{(t)}$, $\overline{w}=\overline{\theta}_G$, and $\overline{w^2}=\overline{\theta^2}_G$
            \State Calculate $\widehat{D}_t = \theta_G^{(t)}-\overline{\theta}_G$
            \If {\texttt{num\_cols}($\widehat{D}$) $>K$}
                \State remove first column of $\widehat{D}$
            \Else 
                \State Append column $\widehat{D}_t$ to $\widehat{D}$
            \EndIf
        \EndIf
    \EndFor
    \State calculate $\Sigma_\text{diag}$ using Equation~\eqref{eq:diagcov}
    \State calculate $\Sigma_\text{l-rank}$ using Equation~\eqref{eq:covl}
    \Statex \textbf{Output: } Approximate posterior $\mathcal{N}(\overline{\theta}_G, \alpha_G(\Sigma_\text{diag} + \Sigma_\text{l-rank}))$ 
    \end{algorithmic}
\end{algorithm*}

The approximate posterior distribution of $\theta_G$ can now be applied for data synthesis. Algorithm~\ref{alg:swagctGAN} outlines the data synthesis procedure using the GACTGAN model. We first define the number of synthetic data rows and its batch size, which implies the iteration of posterior sampling. For instance, if one wants a dataset consisting of 3000 entries and with batch size of 500, there are six iterations. For each iteration, random noise is generated in a predefined batch size, followed by posterior sampling of the weights. The noise is fed to the sampled weights for data synthesis. The process is repeated until the iteration in numbers is finished. The sampling process using different weights effectively indicates exploration in different distribution modes, thereby enhancing the diversity of the generated data~\citep{saatci2017bayesian}.

\begin{algorithm*}[ht]
    \caption{GACTGAN data synthesis}
    \label{alg:swagctGAN}
    \begin{flushleft}
    \textbf{Input: } $n_\text{sample}$, Generator approximate posterior $\mathcal{N}(\overline{\theta}_G, \alpha_G(\Sigma_\text{diag} + \Sigma_\text{l-rank})$ , batch size $M$, number of MC sample $S$
    \end{flushleft}
    \begin{algorithmic}[1]
    \State $T_{s}=\lceil n_\text{sample}/M \rceil$ \Comment{obtain number of iterations for sampling from posterior}
    \State Initiate $\tilde{X}$ as the synthetic data storage
    \For {$t=1\rightarrow T_{s}$}
        \State Sample $z\sim\mathcal{N}(0,\textbf{I})$ with size $M$
        \For {$s=1\rightarrow S$}
            \State Sample posterior $\widetilde{\theta}_G\sim\mathcal{N}\Big(\overline{\theta}_G, \alpha_G(\Sigma_\text{diag} + \Sigma_\text{l-rank})\Big)$
            \State Update batch norm statistics
            \State $\tilde{x} += \frac{1}{S} x^*$ where $x^*$ is obtained by feeding $z$ to $\widetilde{\theta}_G$
        \EndFor
        \State Append row $\tilde{x}$ to $\tilde{X}$
    \EndFor
    \State Transform $\tilde{X}$ to make $\widetilde{\mathcal{D}}$
    \Statex \textbf{Output: } Synthetic dataset $\widetilde{\mathcal{D}}_{1:n_\text{sample}}$
    \end{algorithmic}
\end{algorithm*}

\subsection{Complexity Analysis}

The complexity analysis for CTGAN, BayesCTGAN, and GACTGAN reveals a clear trade-off between computational cost and the benefits of being Bayesian. Note that the analysis is performed with similar generator and discriminator networks for each algorithm.

\textbf{Training complexity. } Let $M$ be the number of mini-batch parameter updates per epoch, and $n_{\text{epochs}}$ be the total number of training epochs, and let $t_{\text{collect}}$ be the epoch at which the SWAG process begins. GACTGAN still maintains good efficiency during training and incurs a small constant overhead per epoch to update its moment matrices ($\mathcal{O}(1)$) for $n_{\text{epochs}} - t_{\text{collect}}$ iterations. Thus, its overall training complexity of $\mathcal{O}(n_{\text{epochs}} \times (M + 1) - t_{\text{collect}})$ remains linear and closely aligned with the baseline CTGAN complexity of $\mathcal{O}(n_{\text{epochs}} \times M)$.  This makes the training process highly scalable despite the added Bayesian machinery.

\textbf{Parameter complexity. } Let $P_G$ denote the number of generator network weights stored. Unlike MCMC-based methods like BayesCTGAN, which must store a large number $S$ of full parameter sets ($\mathcal{O}(S \times P_G)$), GACTGAN employs a more efficient approach by storing only three core components: a mean vector $\overline{\theta}_G$, a diagonal covariance $\Sigma_\text{diag}$, and a low-rank approximation matrix $\Sigma_\text{l-rank}$, resulting in a space complexity of $\mathcal{O}(3 \times P_G)$. This represents substantial memory savings, as $3$ is much smaller than the number of samples $S$ required for good MCMC posterior samples.

\textbf{Sampling complexity. } Finally, the complexity analysis of the sampling shows that the cost of generating the data is application-dependent. For a single sample ($S=1$), the cost of GACTGAN of $\mathcal{O}(T_s)$ is equivalent to CTGAN. However, when generating from multiple posterior samples to enable model averaging, the complexity scales linearly to $\mathcal{O}(T_s \times S)$. This cost is inherent to any Bayesian method that uses an ensemble of models but is justified by the ability to produce diverse outputs and quantify predictive uncertainty, features absent in the CTGAN.

In conclusion, GACTGAN offers a computationally efficient pathway to Bayesian deep learning for generative models. It achieves a favourable balance, introducing only minimal overhead during training and requiring a fixed, low-memory footprint for storing the posterior approximation, all while providing the benefits of a Bayesian approach.

\section{Methodology}
\label{sec:method}

We introduced GACTGAN in Section~\ref{sec:swagctgan} as our first contribution. This section describes the methodology used to support our remaining contributions. We first describe the datasets in Section~\ref{subsec:dataset}. Section~\ref{subsec:exp-setup} then details the experimental framework, including the baseline comparisons and model configurations in Section~\ref{ssec:baselines}. Meanwhile, Section~\ref{ssec:swactgan-impl} presents the implementation of the proposed GACTGAN variants, addressing our second contribution. Finally, Section~\ref{subsec:eval} outlines the evaluation metrics used to assess performance and to examine our third contribution.

\subsection{Dataset}
\label{subsec:dataset}

In this study, we used seven datasets, of which five are census datasets, as shown in Table~\ref{tab:dataset}. Census datasets used are from different countries: UK, Indonesia, Fiji, Canada, and Rwanda, which we obtained from IPUMS International~\citep{ipums}. Census datasets are valuable for benchmarking tabular data synthesis because they capture a range of community characteristics, such as demographics and education. Moreover, these datasets are collected by the National Statistics Offices, which are legally required to ensure that the released data accurately represent the population while also protecting individual privacy~\citep{nowok2016synthpop1,ran2024multiobjective}. Therefore, census datasets are highly relevant as benchmarks for evaluating synthetic data methods. However, one drawback is that census data often lack numeric variables. To address the limitation in terms of numerical attributes, we added two social datasets from the UCI repository, which have also been used in previous research on tabular deep generative models~\citep{xu2019modeling,ma2020vaem,kotelnikov2023tabddpm}. The column details of the datasets can be seen in Table~\ref{tab:columns} in Appendix~\ref{app:data}.

\begin{table}[ht]
    \centering
    \begin{tabular}{*6c}  \toprule
        Acronym & Dataset Name & Year & \#Observations & \#Numerical & \#Categorical \\
         &  &  &  & Variables & Variables\\  \midrule
        UK & UK Census & 1991 & 104267 & 0 & 15\\
        ID & Indonesia Census & 2010 & 177429 & 0 & 13\\
        CA & Canada Census & 2011 & 32149 & 3 & 22 \\
        FI & Fiji Census & 2007 & 84323 & 0 & 19\\
        RW & Rwanda Census & 2012 & 31455 & 0 & 13\\
        AD & Adult & 1994 & 48842 & 5 & 10 \\
        CH & Churn Modelling & N/A & 10000 & 4 & 7 \\ \bottomrule
    \end{tabular}
    \caption{Datasets used in this study and their key characteristics.}
    \label{tab:dataset}
\end{table}

\subsection{Experimental Setup}
\label{subsec:exp-setup}

\subsubsection{Baselines Methods and Configurations}
\label{ssec:baselines}

We compared our proposed methods with several baselines. To answer our main question, we compare this with regular CTGAN as a baseline, which is available as a library in Python\footnote{\url{https://github.com/sdv-dev/CTGAN/blob/main/ctgan}}. The default loss function in the script is Wasserstein; therefore, we add the vanilla loss to align with our study. Furthermore, we also compared our proposed algorithm with BayesCTGAN, an integration of Bayesian GAN developed by~\citep{saatci2017bayesian} with CTGAN\footnote{We adopted the original implementation then changed the optimiser into PSGLD. The original code for pytorch is available in \url{https://github.com/vasiloglou/mltrain-nips-2017/tree/master/ben\_athiwaratkun/pytorch-bayesgan}}. However, instead of using SGHMC, we sampled the parameters using preconditioned stochastic gradient Langevin dynamics (PSGLD)~\citep{li2015preconditioned}\footnote{The original code is available in \url{https://github.com/automl/pybnn/blob/master/pybnn/sampler/preconditioned\_sgld.py}}. 

Table~\ref{tab:config} shows the algorithm parameters and their settings as used in the study. For fair comparison, we put a similar basic configuration for all datasets without fine-tuning, mainly adopted from~\cite{xu2019modeling}. We used two hidden layers of size 256, as recommended in previous studies as one of the combinations of hidden layers~\citep{kotelnikov2023tabddpm, gorishniy2023revisiting}. We also adapted the PacGAN framework~\citep{lin2018pacgan}, which uses several discriminator samples to improve the sample quality. We used Adam as the optimiser~\citep{kingma2017adam}, except for BayesCTGAN which uses PSGLD with different prior variances ($\sigma^2_\text{prior}$). We only sweep the $\sigma^2_\text{prior}$ to make sure that the difference only come from the Bayesian setting.

\subsubsection{GACTGAN variants implementation.}
\label{ssec:swactgan-impl}

We used different variants of GACTGAN as our proposed method. The first variant uses SWA without any posterior approximation, which we denote as the averaging CTGAN or ACTGAN. In GACTGAN, we investigated performance in diagonal and low-rank covariance settings to answer our \textbf{second contribution}. Further, we tested different covariance rank settings, including diagonal (D), 30, 100, and 150. For clarity, we label each version according to its covariance rank, for example, GACTGAN (D), GACTGAN (30), and so forth. The fourth part of Table~\ref{tab:config} summarises the configurations for GACTGAN.

We acknowledge that GACTGAN requires tuning that can be done during the synthesis phase without modifying the training parameters, which becomes the advantage over CTGAN and BayesCTGAN. In the synthesis phase, we used different covariance scales $\alpha_G$, in which the findings of~\cite{maddox2019simple} showed that it is recommended to use a scale below 1. 

\begin{table}[t]
    \centering
    \begin{tabular}{p{0.25\linewidth}p{0.3\linewidth}p{0.3\linewidth}}\toprule
        Model & Parameter & Size \\ \midrule
        \multirow{6}{*}{CTGAN} & layer size & (256, 256) \\
         & Noise dimensions & 128 \\
         & Dropout (for $D$) & 0.5 \\
         & Number of epochs & 200 \\
         & Batch size & 500 \\
         & Pac & 10 \\ \midrule
        \multirow{4}{*}{Optimiser} & Algorithm & Adam \\
         & Learning rate & $2\times 10^{-4}$ \\
         & Weight Decay & $1\times 10^{-6}$ \\
         & Coefficients & (0.5, 0.9) \\ \midrule
        \multirow{3}{*}{BayesCTGAN} & Preconditioning decay & 0.99 \\
         & $\sigma^2_{prior}$ & 0.01, 1, 10 \\ 
         & \#MCMC samples & 20 \\ \midrule
        \multirow{3}{*}{GACTGAN} & $\sigma^2_{prior}$ & 1 \\
         & $K$ & 0 (Diagonal), 30, 100, 150 \\ 
         & $\alpha_G$ & 0 (ACTGAN), 0.25, 0.5, 1.0 \\ \bottomrule
    \end{tabular}
    \caption{Algorithm parameters and their settings as used in this study.}
    \label{tab:config}
\end{table}

\subsection{Evaluation Method}
\label{subsec:eval}

\subsubsection{Qualitative and Quantitative Evaluations}

We evaluated the performance of the synthetic data produced using different approaches. For starters, we provided an evaluation visually to assess the behaviour of the synthetic data generated by the models compared to the real data. For continuous variables, we used the absolute difference in correlation to investigate the relationship between variables and histogram to investigate the distribution. On the other hand, we visualised the categorical variables using a bar chart based on the cross-tabulation results.

We also performed quantitative evaluations based on their utility and disclosure risk values. We focus on the narrow utility measures~\citep{taub2019the}, which include the user's perspective on what they want to use the data for. Narrow measures can be calculated using two approaches: ratio of counts (ROC) and confidence interval overlap (CIO). ROC compares cell values within frequency tables and cross-tabulations between synthetic and original data by calculating the ratio between the smaller of any cell count pairs and the larger of the pair~\citep{little2021generative}. CIO measures the performance of synthetic data in statistical inference using a statistical model, such as the coefficients of the logistic regression model~\citep{ran2024multiobjective}.

Disclosure risk values, on the other hand, can be found using marginal target correct attribution probability (TCAP) values, as seen in~\eqref{eq:risk}. These show how well an adversary can determine sensitive variables from fake data, as long as they know some of the population data~\citep{little2021generative}. Considering the baseline, we truncated the rescaled TCAP values to zero to obtain the risk under the assumption that TCAP values below zero already have low risk. Truncating the values also gives easier interpretation and comparison. The details of the measurements can be seen in~\citep{taub2019the,little2021generative,taub2019the,ran2024multiobjective}.

\begin{equation}
    \label{eq:risk}
    R = \max\{0,\frac{TCAP - WEAP}{1-WEAP}\}
\end{equation}

Since many combinations of models are to be investigated, we use an aggregate score to select the best model based on the research questions we want to answer, which accounts for the utility and risk. The selection score (SS) is calculated in Equation~\eqref{eq:score}.

\begin{equation}
    \label{eq:score}
    SS = (\phi\times U)+[(1-\phi)\times(1-R)]
\end{equation}

$\phi$ is the weight of the utility, which is [0,1]. Increasing $\phi$ will focus the model selection on utility and vice versa. In this study, we used $\phi=0.75$ as we are looking for a model with balanced utility and risk. For each simulation, we generated five different datasets from different seeds using the trained models. We then calculated utility and risk, followed by mean calculations across the different generated datasets. 

\subsubsection{Pareto Front}

In multi-objective optimisation, the Pareto front provides a principled way to analyse trade-offs among competing objectives and to identify solutions that best align with a decision maker's preferences~\citep{Mirjalili2020multi}. The concept is grounded in \textbf{Pareto dominance}~\citep{censor1977pareto}, which formalises when one solution is considered superior to another. Specifically, a solution \( s_1 \) is said to \textit{dominate} another solution \( s_2 \) if and only if:
\begin{enumerate}
    \item \( s_1 \) is at least as good as \( s_2 \) in \textbf{all} objectives, and 
    \item \( s_1 \) is better than \( s_2 \) in at least \textbf{one} objective.
\end{enumerate}

A solution is considered \textbf{Pareto optimal} if it is not dominated by any other feasible solution. Unlike single-objective optimisation problems, which typically admit a single best solution, multi-objective problems generally yield a set of Pareto-optimal solutions~\citep{Allmendinger2016navigation}. This arises because improvements in one objective often lead to degradation in another, resulting in inherent trade-offs rather than a single optimal outcome. The collection of Pareto-optimal solutions constitutes the \textbf{Pareto front}, which serves as a pool of candidate solutions from which decision makers can select according to their preferences. In this work, we use the Pareto front to analyse the trade-off between utility, which is to be maximised, and risk, which is to be minimised. Visualising these objectives on a two-dimensional plane provides an intuitive and effective means of identifying suitable solutions~\citep{Little2024synthetic}.


\section{Results}
\label{sec:results}

This section disseminates the results obtained from our experiments. Initially, in Section~\ref{subsec:descriptive}, we provided descriptive statistics related to both synthetic data and original data. Furthermore, we delved into the primary quantitative findings and engaged in an in-depth discussion of these results in Section~\ref{subsec:quant}. We also added additional analysis of different GACTGAN parameters in Section~\ref{subsec:consider}. Lastly, Section~\ref{subsec:ablation} provides insight into how our proposed model performs in different scenarios.

\subsection{Descriptive Statistics Comparison}
\label{subsec:descriptive}

\begin{figure}[ht]
    \centering
    \includegraphics[width=1\textwidth]{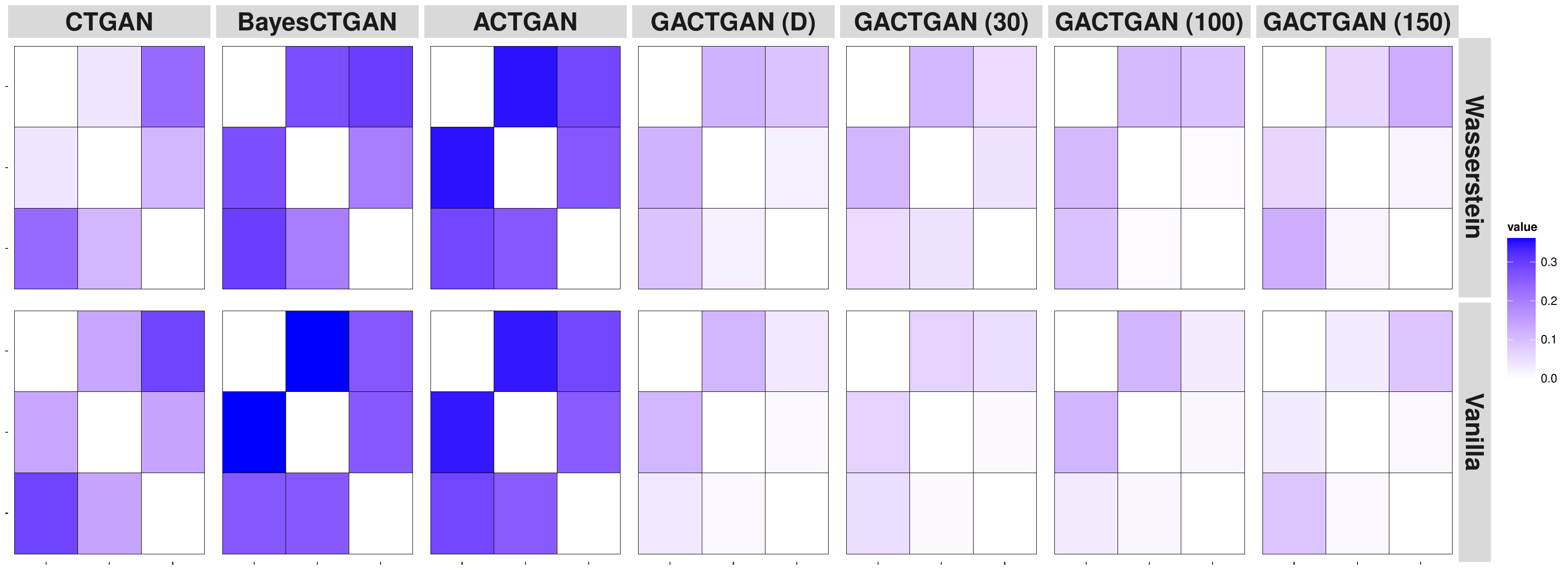}        
    \caption{Heatmap visualisation of correlation difference for the CA dataset across different generative models (columns) and loss functions (rows). The difference is computed by subtracting the real data correlation from the synthetic data correlation. Each inner $3 \times 3$ grid displays the pairwise correlation differences between synthetic and real data for the dataset's continuous variables. Lighter tiles indicate minimal deviation from the real data (better preservation of multivariate dependencies), while darker blue tiles highlight otherwise. GACTGAN variants (right columns) predominantly exhibit lighter tones, demonstrating superior preservation of correlation structures compared to baselines like CTGAN and BayesCTGAN.}
    \label{fig:quali-corr}
\end{figure}

\begin{figure}
    \centering
    \subfloat[UK-House ownership among females]{
      \includegraphics[width=\textwidth]{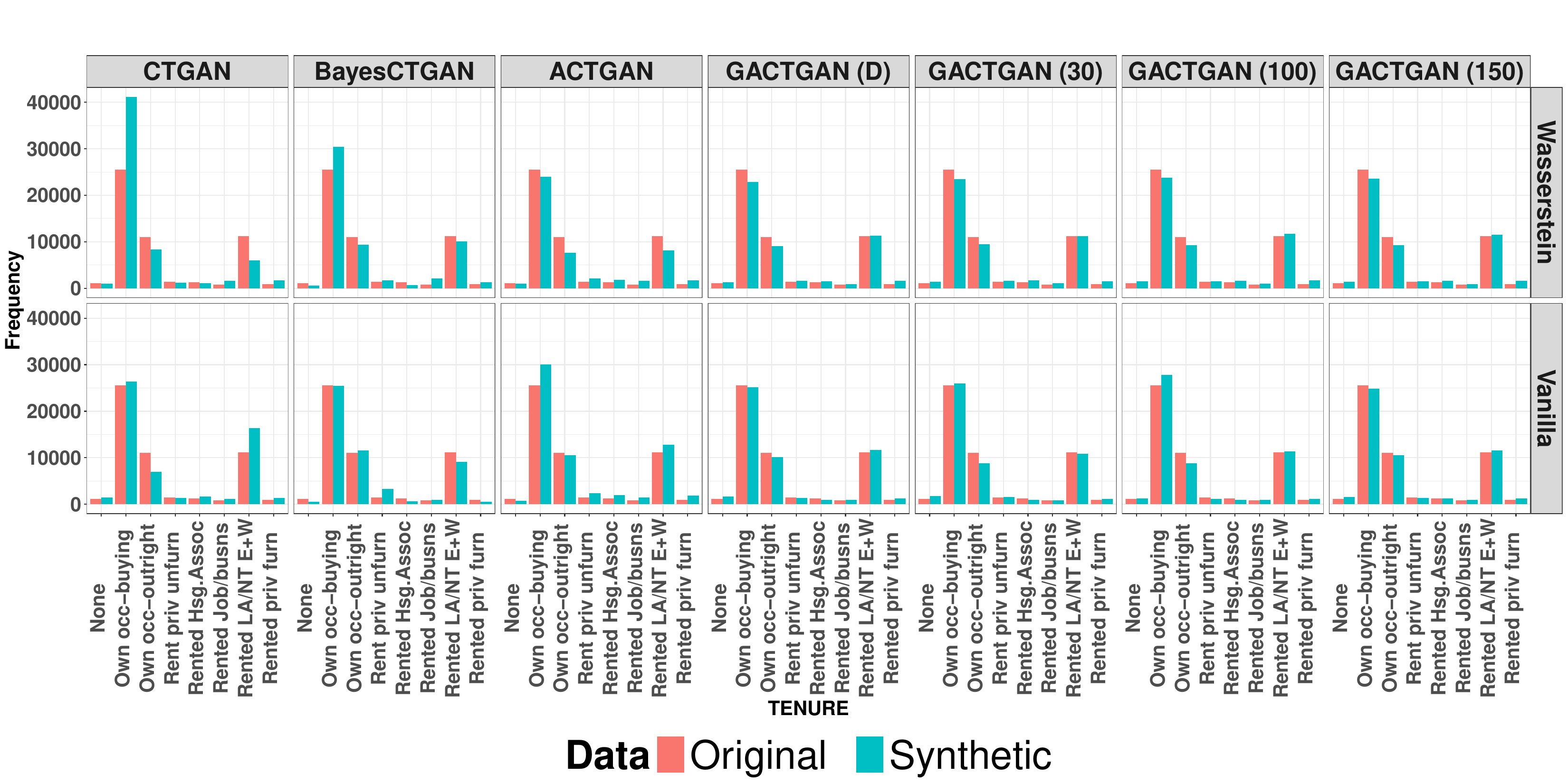}
    } \\ 
    \subfloat[CH-credit score]{
      \includegraphics[width=\textwidth]{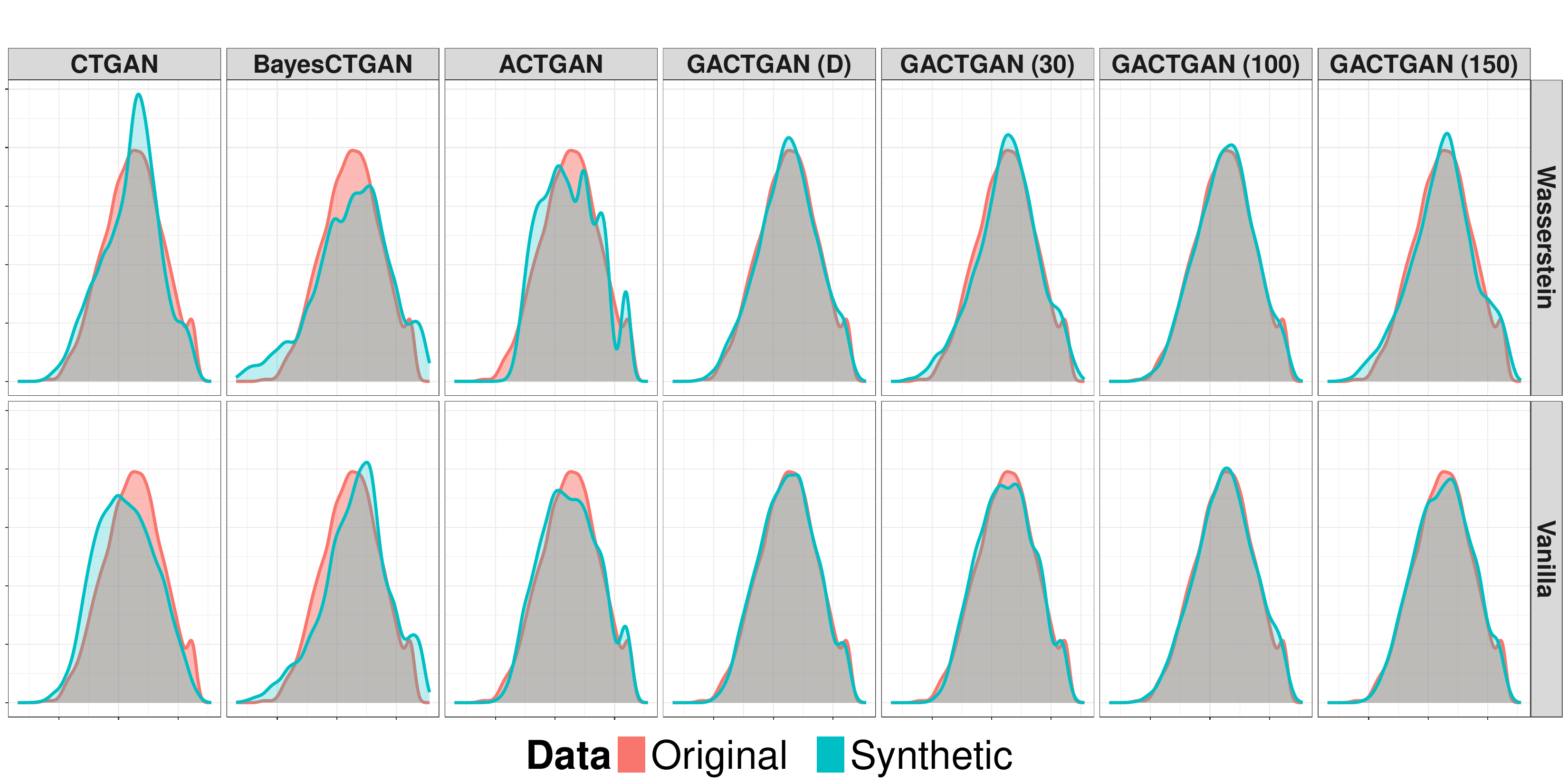}
    } \\
    \caption{Descriptive statistics of synthetic and original data: (a) and (b) represent cross-tabulation of house ownership among different sex in the UK, and (c) and (d) represent the distribution of estimated salary and credit score in CH, respectively. Each subplot compares the synthetic data generated by different methods (CTGAN, BayesCTGAN, ACTGAN, GACTGAN with varying parameters) to the original data, highlighting the alignment between the synthetic and original distributions across different demographics and variables. The top and bottom part of each subfigure represent Wasserstein and vanilla loss.}
    \label{fig:quali-des}
\end{figure}

Before going into quantitative evaluation, we attempted to visually compare the synthetic data with respect to the original counterparts. We analysed continuous variables using correlation matrices and histograms and categorical data output using cross-tabulations between two variables. Considering page limitation, we only provide a few examples from different datasets, as seen in Figures~\ref{fig:quali-corr} and~\ref{fig:quali-des}. In each subfigure, the top row corresponds to a Wasserstein loss, while the bottom row represents a vanilla loss. 

Figure~\ref{fig:quali-corr} presents the correlation differences between synthetic and real data in the CA dataset. The difference is computed by subtracting the real data correlation from the synthetic data correlation, with brighter colours (lighter purple) indicating better ability to preserve correlation between numerical variables and darker colours (blue) showing otherwise. In the figure, each method is represented by a $3 \times 3$ heatmap corresponding to the pairwise correlations of the three continuous variables in the dataset. The results show that GACTGAN consistently outperforms CTGAN and BayesCTGAN. These configurations achieve the lowest correlation differences, indicating closer correlation of the real data. In contrast, CTGAN, BayesCTGAN, and ACTGAN exhibit higher deviations, with darker colours indicating their inability to preserve correlations. Therefore, it can be concluded that GACTGAN performed better in preserving correlations.

Figure~\ref{fig:quali-des} reveals the superior performance of GACTGAN in capturing the distribution within categorical and continuous variables. In Figure~\ref{fig:quali-des}a, GACTGAN accurately captures the frequency of category ownership in the UK between men, which is indicated from the blue chart to the red in GACTGAN. In contrast, CTGAN, BayesCTGAN, and ACTGAN exhibit over- and underestimation in some categories, especially in minor categories, failing to replicate the real data's patterns effectively. In Figure~\ref{fig:quali-des}b, GACTGAN continues to excel in learning the distributions of the credit score in the CH dataset. Similarly, for the credit score, GACTGAN achieves high fidelity, accurately replicating the original data distribution. However, using lower covariance ranks, such as diagonal, show reduced performance, underscoring the importance of the rank. In comparison, CTGAN, BayesCTGAN, and ACTGAN fail to capture the finer details of the distribution, particularly in the peak and tail. In general, GACTGAN demonstrates its robustness and adaptability across various variables and datasets, significantly outperforming other methods to preserve categorical and continuous data distributions. These results highlight the importance of leveraging SWAG's posterior approximation in the CTGAN generator and carefully tuning covariance ranks for optimal synthetic data generation.

In summary, GACTGAN with higher parameter values (100 and 150) provides the best overall performance, preserving the original data characteristics across all variables and datasets. Other methods show less consistent results, making GACTGAN the most reliable choice for generating high-quality synthetic data. We also provided additional descriptive statistics in the Appendix~\ref{app:adddesc}. The next subsection shows our evaluation results from a quantitative perspective, which completes our findings.

\subsection{Quantitative Results}
\label{subsec:quant}

After observing that our proposed method performed better in descriptive statistics, this subsection discusses our main goal, the quantitative investigation of whether GACTGAN could improve CTGAN in terms of utility and risk. We did the simulation results quantitatively using the evaluation measurements mentioned in Section 4. We run the models based on the considered hyperparameters, followed by a selection based on the maximum selection score. As an initial ablation study, we also placed SWA (denoted as ACTGAN) and SWAG-diagonal (denoted as GACTGAN (D)) as the special case of SWAG. We also compared different covariance ranks of GACTGAN as denoted inparentheses.

\begin{figure}
    \centering
    \includegraphics[width=1\textwidth]{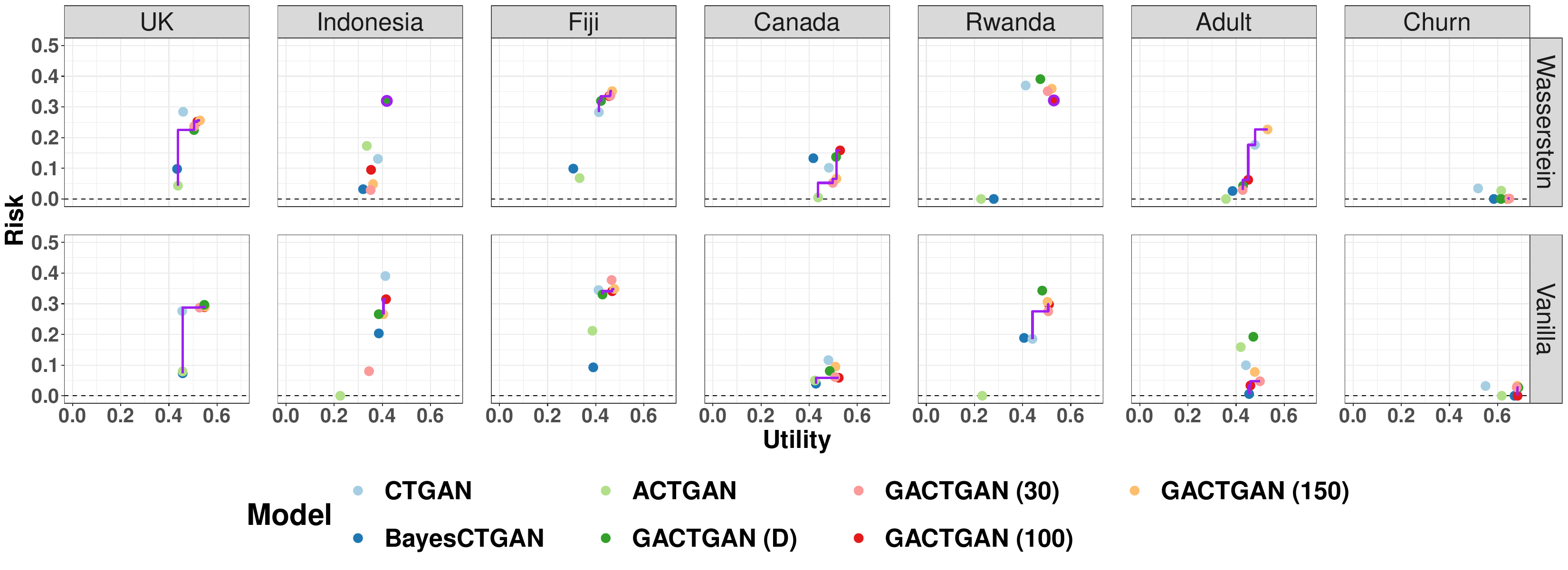}        
    \caption{R-U map of CTGAN, Bayesian GAN, and GACTGAN. The purple lines indicated the solution candidates based on Pareto front.}
    \label{fig:rumap-final}
\end{figure}

Figure~\ref{fig:rumap-final} showed the simulation results of CTGAN, BayesCTGAN, and GACTGAN divided by loss functions and visualised on the R-U map with the obtained Pareto front to select the most optimal solution candidates. We only consider algorithms that have a utility larger than 0.4 to be included in the candidate because a dataset with utility lower than 0.4 cannot be properly used. For example, although BayesCTGAN in RW can be a solution candidate based on the Pareto front, it is excluded because the utility is less than 0.4. 

In general, Figure~\ref{fig:rumap-final} shows that BayesCTGAN and SWA-CTGAN performed worse compared to other algorithms. It can be seen from the figure that SWA-CTGAN has the lowest utility in most datasets both in vanilla and in Wasserstein loss, although they are sometimes chosen as a solution in some datasets. Moreover, CTGAN was found to perform better than both algorithms while performing worse in CH. Therefore, we do not suggest using weight averaging in the generator for tabular data synthesis, except probably for small datasets. On the other hand, the Pareto front, visualised in the purple line, showed that GACTGAN became the primary candidate solution for tabular data synthesis in all datasets, with CTGAN joined in some datasets.

In detail, at the top of Figure~\ref{fig:rumap-final}, it is evident that in the Wasserstein distance, GACTGAN could improve upon CTGAN. The claim can be seen from the improvement of utility in most datasets, such as FI and AD, by GACTGAN with rank 150. Moreover, our interesting finding is that in the UK, CA, RW, and CH, the synthetic data generated from GACTGAN seem to have higher utility and lower disclosure risk, which is ideal for tabular data synthesis. However, in the ID dataset, the utility of the SWAG dataset is slightly lower than that of CTGAN, but the risk becomes significantly lower, but they are not considered a solution because of the utility cut-off point. The same thing happened for GACTGAN in AD except for rank 150, but they are preferred solution.

On the other hand, at the bottom of Figure~\ref{fig:rumap-final}, where the models use vanilla loss, there is a pattern similar to that of the top figure. In the UK and FI, GACTGAN has a larger utility in general with a slight difference in risk with CTGAN. In CA, AD, and CH, the ideal scenario of synthetic data protection occurred. In RW, the utility and risk values increase compared to CTGAN, which commonly occurs in tabular data synthesis. In the ID dataset, GACTGAN has a lower risk than CTGAN with the same level of utility.

In summary, GACTGAN could improve CTGAN by producing a more useful synthetic dataset with lower disclosure risk. Considering the focus on GACTGAN, it can be seen that using different covariance ranks gives quite close results in the datasets, whereas using the covariance approximation showed improvements compared to the diagonal. 


\subsection{Utility, Loss, and Rank Consideration in GACTGAN}
\label{subsec:consider}

Table~\ref{tab:swag} presents the utility and risk gains (\%) for different covariance levels, utility weights (\(\phi\)), and loss functions when combining SWAG with CTGAN. In this updated context, \(\phi = 1\) represents \textbf{full utility weight}, while \(\phi = 0.75\) indicates \textbf{partial utility weight}. The results highlight how different configurations affect the balance between utility and risk, offering insights for optimising performance.

For \(\phi = 1\), the vanilla loss function achieves its highest utility gain of 6.46\% at a covariance rank of 150, with a moderate risk of 4.08\%. Lower ranks, such as diagonal rank, show reduced utility (4.63\%) but maintain a much lower risk (1.30\%). The Wasserstein loss function also performs well with \(\phi = 1\), achieving its highest utility of 6.95\% at rank 150, but at a slightly higher risk of 4.29\%. However, in lower ranks, such as the diagonal rank, the Wasserstein loss sacrifices utility (3.63\%) to achieve its lowest risk of 5.15\%.

When using \(\phi = 0.75\), the trends change slightly. The Vanilla loss function still favours higher covariance ranks, with the highest utility gain of 6.18\% achieved at rank 150. However, the risk remains relatively low at 1.29\%. The Wasserstein loss function demonstrates a strong balance of utility and risk at rank 100, where utility reaches 6.08\% and risk is reduced to 2.69\%. Lower covariance ranks, such as the diagonal, again show reduced utility but maintain better risk levels for certain configurations.

Based on observations, full utility weight is recommended for tasks that prioritise utility, as it consistently yields higher utility gains across different configurations. In contrast, a partial utility weight may be more suitable for risk-sensitive applications due to its better control over risk reduction. The Wasserstein loss function outperforms the vanilla loss function in many cases, particularly when prioritising a balance between utility and risk. As for the covariance rank, higher ranks (for example, 150) generally perform better to maximise utility, while moderate ranks (e.g., 100) strike a balance between utility and risk, making them a robust choice for most applications. However, the computational cost is prohibitively expensive if the rank is too large, which should also be taken into consideration when building GACTGAN.

\begin{table}[ht]
    \centering
    \resizebox{\textwidth}{!}{
    \begin{tabular}{ccccccccc}
        \toprule
        \multirow{3}{*}{Covariance Rank} & \multicolumn{4}{c}{$\phi$ = 0.75} & \multicolumn{4}{c}{$\phi$ = 1} \\ 
        \cmidrule(lr){2-5} \cmidrule(lr){6-9}
         & \multicolumn{2}{c}{Vanilla} & \multicolumn{2}{c}{Wasserstein} & \multicolumn{2}{c}{Vanilla} & \multicolumn{2}{c}{Wasserstein} \\ 
         \cmidrule(lr){2-3} \cmidrule(lr){4-5} \cmidrule(lr){6-7} \cmidrule(lr){8-9}
         & Utility ($\uparrow$) & Risk ($\downarrow$) & Utility ($\uparrow$) & Risk ($\downarrow$) & Utility ($\uparrow$) & Risk ($\downarrow$) & Utility ($\uparrow$) & Risk ($\downarrow$) \\ 
         \midrule
        Diagonal & 4.21 & 1.30 & 3.24 & 0.78 & 4.63 & 4.29 & 3.63 & 5.15 \\
        30 & 4.86 & -4.12 & 3.57 & -4.89 & 6.18 & 1.29 & 5.73 & 4.59 \\
        100 & 5.93 & -1.57 & 4.74 & -2.22 & 6.53 & 1.57 & 6.08 & 2.69 \\
        150 & 5.89 & -0.41 & 6.03 & -1.02 & 6.46 & 4.08 & 6.95 & 4.29 \\
        \bottomrule
    \end{tabular}
    }
    \caption{Average utility and risk gain (in \%) of GACTGAN with different covariance ranks using partial utility weight ($\phi=0.75$) and full utility weight ($\phi=1$) in the model selection. The gain is calculated using the difference of average utility and risk of GACTGAN w.r.t. CTGAN.}
    \label{tab:swag}
\end{table}

\subsection{Ablation Study}
\label{subsec:ablation}

\begin{figure}
    \centering
    \includegraphics[width=1\textwidth]{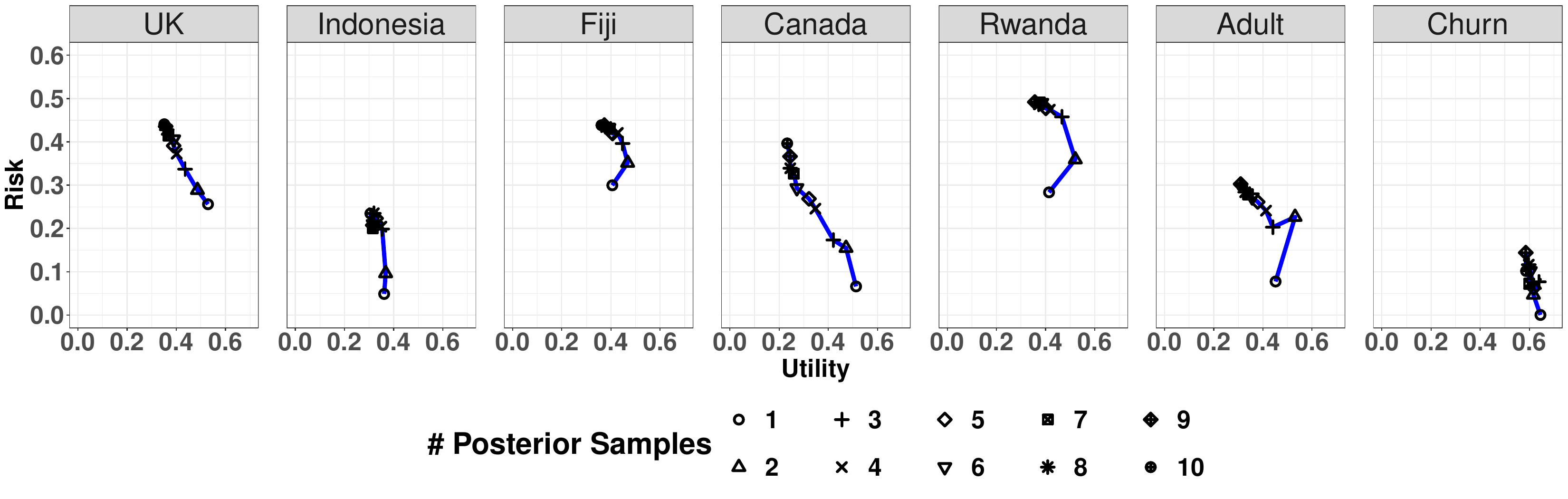}
    \caption{Utility-Risk map of GACTGAN based on number of posterior samples. The blue line showed the trend between each number of samples.}
    \label{fig:rumap-abl2}
\end{figure}

The beginning of our quantitative results is considered as an ablation study because of the different models to use, namely ACTGAN and GACTGAN (D). This subsection provides additional ablation study based on the number of posterior samples for BMA on different data. Figure~\ref{fig:rumap-abl2} shows the R-U map of the GACTGAN synthesis results using the difference number of posterior samples. Based on the figure, an inverse diagonal trend can be seen across the increment of number of posterior samples. The finding indicates that the use of large posterior samples of BMA for tabular data synthesis in CTGAN is not recommended because it tends to generate less useful data with higher disclosure risk, which should be avoided in tabular data synthesis. On the other hand, in some datasets, such as RW and AD, using two samples could produce data that has higher utility but with increased risk. Based on~\citep{saatci2017bayesian}, the sampled weight from $G$ is used to explore different modes of distribution, instead of model averaging. Thus, we recommend trying between one or two posterior samples when performing data synthesis.

\section{Discussion}
\label{sec:discussion}

Our comprehensive evaluation across seven diverse tabular datasets demonstrates that GACTGAN consistently outperforms existing benchmarks, including standard CTGAN, its Bayesian variant (BayesCTGAN), and a generator using Stochastic Weight Averaging (SWA), across both descriptive and quantitative metrics. This section interprets these findings, situates them within the broader literature on synthetic data and Bayesian deep learning, and offers a critical analysis of the implications, strengths, and limitations of our approach.

\subsection{Interpretation of Key Findings}

The superior performance of GACTGAN can be attributed to its enhanced ability to capture the complex, multi-modal distributions typical of tabular data. The SWAG posterior provides a more nuanced and robust approximation of the generator's parameter distribution compared to point estimates (CTGAN), mean-field variational inference (BayesCTGAN), or a simple average of weights (SWA). This allows GACTGAN to generate a richer variety of plausible data points, which directly translates to higher utility, as evidenced by its improved correlation preservation (Figure~\ref{fig:quali-corr}) and distributional fidelity (Figure~\ref{fig:quali-des}).

A crucial finding is the critical role of the covariance rank within the SWAG approximation. Our results (Table~\ref{tab:swag}) indicate that a low-rank approximation (e.g., rank 30) often underperforms, while a full-rank (diagonal) approximation, while computationally cheaper, fails to capture parameter correlations, leading to suboptimal results. The highest utility gains were consistently achieved with higher ranks (100-150), which effectively model dependencies between parameters in the generator network. This aligns with the understanding that capturing these correlations is essential to generate coherent and realistic data records~\citep{maddox2019simple}. However, this comes with a non-trivial computational cost, presenting a practical trade-off for practitioners.

Furthermore, the ablation study on the number of posterior samples for Bayesian Model Averaging (BMA) (Figure~\ref{fig:rumap-abl2}) yielded a counter-intuitive yet insightful result: using more than one or two samples often degraded performance. This suggests that for the task of data synthesis--where the goal is to produce a single high-quality dataset--extensive sampling from the posterior may lead to ``averaging out'' the distinct high-quality modes that a single good sample can capture. This finding nuances the typical BMA paradigm and suggests that in generative tasks, exploring a few high-probability modes might be more effective than attempting to average over the entire posterior.

\subsection{Comparison with Prior Work}
Our findings confirm and challenge existing narratives in the literature. The consistent underperformance of BayesCTGAN and SWA-CTGAN (ACTGAN) reinforces the conclusions of~\citet{xu2019modeling} and others that naively applying Bayesian methods to GAN is fraught with difficulty. The training instability and mode collapse issues inherent in GAN are often exacerbated by simplistic variational approximations. GACTGAN addresses this by employing SWAG, a method lauded for its stability and accuracy in capturing posterior distributions in deep networks~\citep{maddox2019simple}, and successfully adapts it to the adversarial training setting of a GAN.

The result that the Wasserstein loss generally outperformed the vanilla loss in terms of utility (Table~\ref{tab:swag}) is consistent with the established benefits of the Wasserstein distance for GAN training, such as improved stability and more meaningful loss gradients~\citep{arjovsky2017wasserstein1}. However, our work adds a new layer to this understanding: the choice of loss function also interacts significantly with the Bayesian approximation method. The Wasserstein loss's smoother landscape likely provides a more stable base over which SWAG can construct an accurate posterior, leading to more reliable improvements.

Our work also contributes to the ongoing discussion on the dilemma between data utility and disclosure risk~\citep{taub2019the,Little2024synthetic}. While GACTGAN primarily optimises utility, the synthetic data was produced without a proportional increase in disclosure risk, and in some cases even reduced it (e.g., in the CA dataset with Wasserstein loss). This suggests that by producing a more accurate and less ``memorised'' representation of the underlying data distribution, a better Bayesian generator can inherently mitigate some privacy risks.

\subsection{Limitations}

Despite its promising results, this study has several limitations that should be acknowledged. First, the computational overhead of GACTGAN is significant. Training SWAG requires maintaining a running estimate of the first and second moments of the weights over multiple epochs, and generating samples involves forward passes through multiple networks drawn from the posterior. This cost may be prohibitive for very large datasets or models, and future work should focus on developing more computationally efficient approximations tailored for generative models. Moreover, future research may also explore other posterior approximations (e.g., Variational Inference~\citep{blundell2015weight,zhang2018noisy}, Monte Carlo Dropout~\citep{gal2016dropout}) and their integration within different generative architectures.

Second, while our evaluation is comprehensive, it is constrained to the CTGAN architecture and specific datasets. The generalisability of SWAG's benefits to other state-of-the-art tabular generators (e.g., TabDDPM~\citep{kotelnikov2023tabddpm}, TVAE~\citep{xu2019modeling}, and Flow matching~\citep{guzman-cordero2025exponential,nasution2026flowmatchingtabulardata}) as well as integration with other generator architectures (e.g. Transformers-based model) remains an open question. It is plausible that the advantages of a Bayesian generator would be even more pronounced in a more powerful base model.

Finally, the optimal configuration of GACTGAN (rank, number of samples, $\phi$) appears to be dataset-dependent. Although we provide general guidelines, the need for per-dataset tuning somewhat diminishes the ``out-of-the-box'' advantage. Developing methods for automated configuration or demonstrating more robust default settings would enhance the practical adoption of this technique.

\section{Concluding Remarks}
\label{sec:concl}

This study introduced GACTGAN, a novel Bayesian extension of the CTGAN framework that leverages Stochastic Weight Averaging-Gaussian (SWAG) for posterior approximation in the generator. We evaluated our proposed algorithm against CTGAN, BayesCTGAN, and ACTGAN across multiple datasets using utility and disclosure risk metrics. Across datasets, GACTGAN produced synthetic samples that more closely matched the empirical properties of the real data, with improvements observed for both categorical and continuous features. These findings highlight SWAG-style posterior approximations as a practical mechanism for improving tabular data synthesis. A consistent empirical finding is that richer covariance structure in the SWAG posterior matters. Higher covariance ranks, particularly around 100, tended to provide the most reliable gains, primarily when combined with the Wasserstein objective. In addition, the results suggest that sampling overhead can often be kept modest, since a small number of posterior samples may already deliver most of the utility benefits. Overall, these results support the use of scalable posterior approximations as a viable route to improving the performance of tabular generative models, especially CTGAN.

\subsection*{Data and Code Availability}
The implementation code of this paper is available in~\url{https://github.com/rulnasution/ctgan-bayes/}.

\subsection*{Acknowledgements}
B.I. Nasution is supported by the Indonesia Endowment Fund for Education Agency (LPDP) scholarship.

\subsection*{Declarations}
The authors declare no conflict of interest.

\bibliographystyle{sn-chicago}
\bibliography{sn-bib}

\newpage
\appendix

\section{Dataset Details}
\label{app:data}

Table~\ref{tab:columns} describes the details of the dataset we used. We differentiate the use of age in census and noncensus datasets. We treated age as a categorical variable in the census dataset, following previous studies~\citep{ran2024multiobjective}. In contrast, we considered age as a numerical variable in the AD and CH datasets. Furthermore, we limited the region to the West Midlands area in the UK dataset. On the other hand, we only used the Deli Serdang municipality area in the ID dataset (code 1212 or 012012 in IPUMS). The details of the variables used to calculate the evaluation measurements can be seen in the code.

\begin{table}[ht]
\caption{Details of columns used in this study.}
    \centering
    \begin{tabular}{ccp{0.15\linewidth}p{0.5\linewidth}}\toprule
        Dataset & Year & \multicolumn{1}{c}{Continuous columns} & \multicolumn{1}{c}{Categorical Columns} \\ \midrule
        UK & 1991 & - & AREAP, AGE, COBIRTH, ECONPRIM, ETHGROUP, FAMTYPE, HOURS, LTILL, MSTATUS, QUALNUM, RELAT, SEX, SOCLASS, TRANWORK, TENURE \\ 
        ID & 2010 & - & OWNERSHIP, LANDOWN, AGE, RELATE, SEX, MARST, HOMEFEM, HOMEMALE, RELIGION, SCHOOL, LIT, EDATTAIND, DISABLED \\
        FI & 2007 & - & PROV, TENURE, RELATE, SEX, AGE, ETHNIC, MARST, RELIGION, BPLPROV, RESPROV, RESSTAT, SCHOOL, EDATTAIN, TRAVEL, WORKTYPE, OCC1, IND2, CLASSWKR, MIG5YR \\
        RW & 2012 & - & AGE, STATUS, SEX, URBAN, REGBTH, WKSECTOR, MARST, NSPOUSE, CLASSWK, OWNERSH, DISAB2, DISAB1, EDCERT, RELATE, RELIG, OCC, HINS, NATION, LIT, IND1, BPL \\
        AD & - & age, fnlwgt, capital-gain, capital-loss, hours-per-week & workclass, education, education-num, marital-status, occupation, relationship, race, sex, native-country, income \\
        CH & - & CreditScore, Age, Balance, EstimatedSalary & Geography, Gender, Tenure, NumOfProducts, HasCrCard, IsActiveMember, Exited\\ \bottomrule
    \end{tabular}
    \label{tab:columns}
\end{table}

\section{Additional Descriptive Statistics}
\label{app:adddesc}
\begin{sidewaysfigure}
    \centering
    \subfloat[RW-Working sector in status=1]{
      \includegraphics[width=0.5\textwidth]{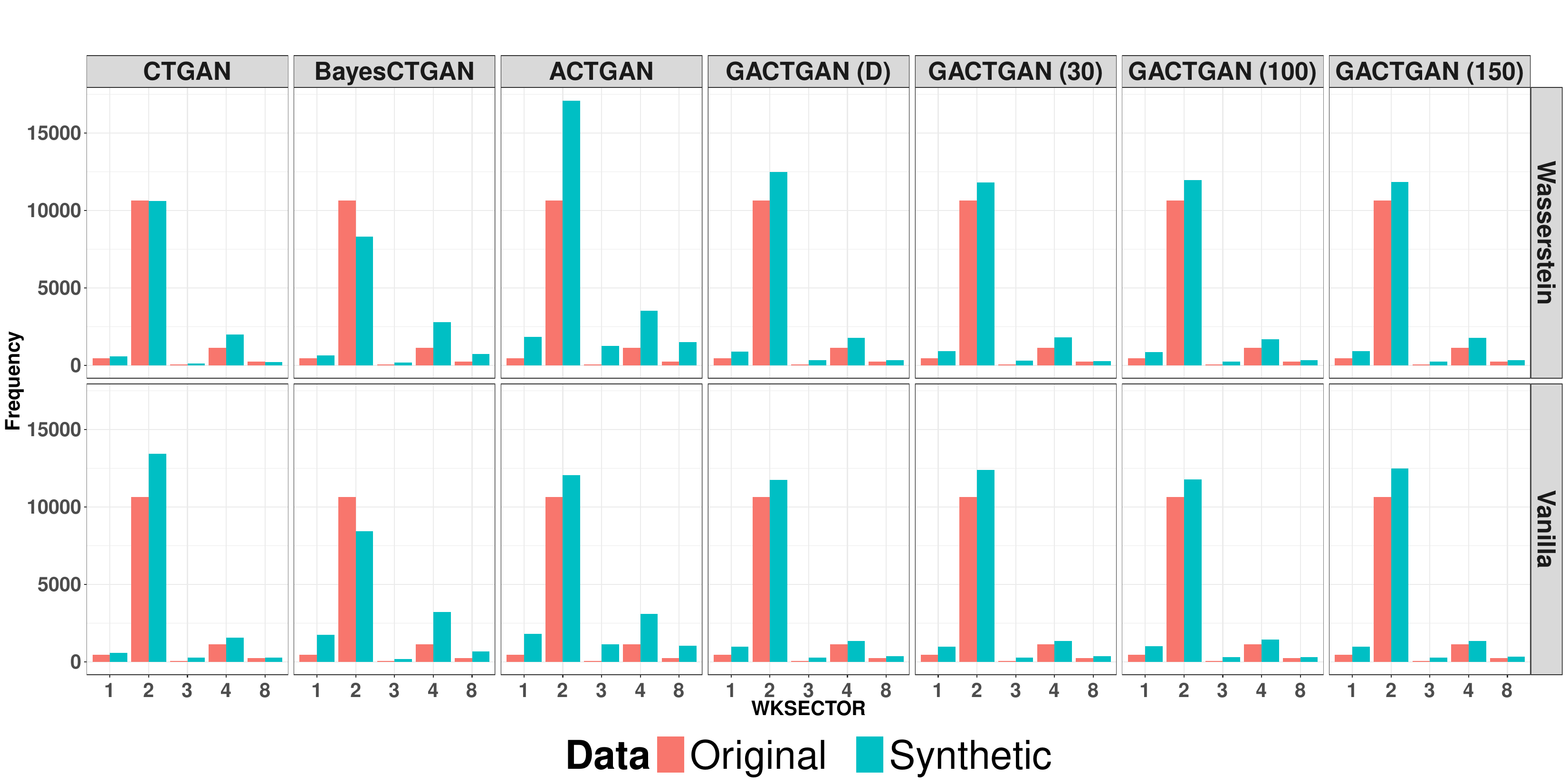}
    } 
    \subfloat[RW-Working sector in status=2]{
      \includegraphics[width=0.5\textwidth]{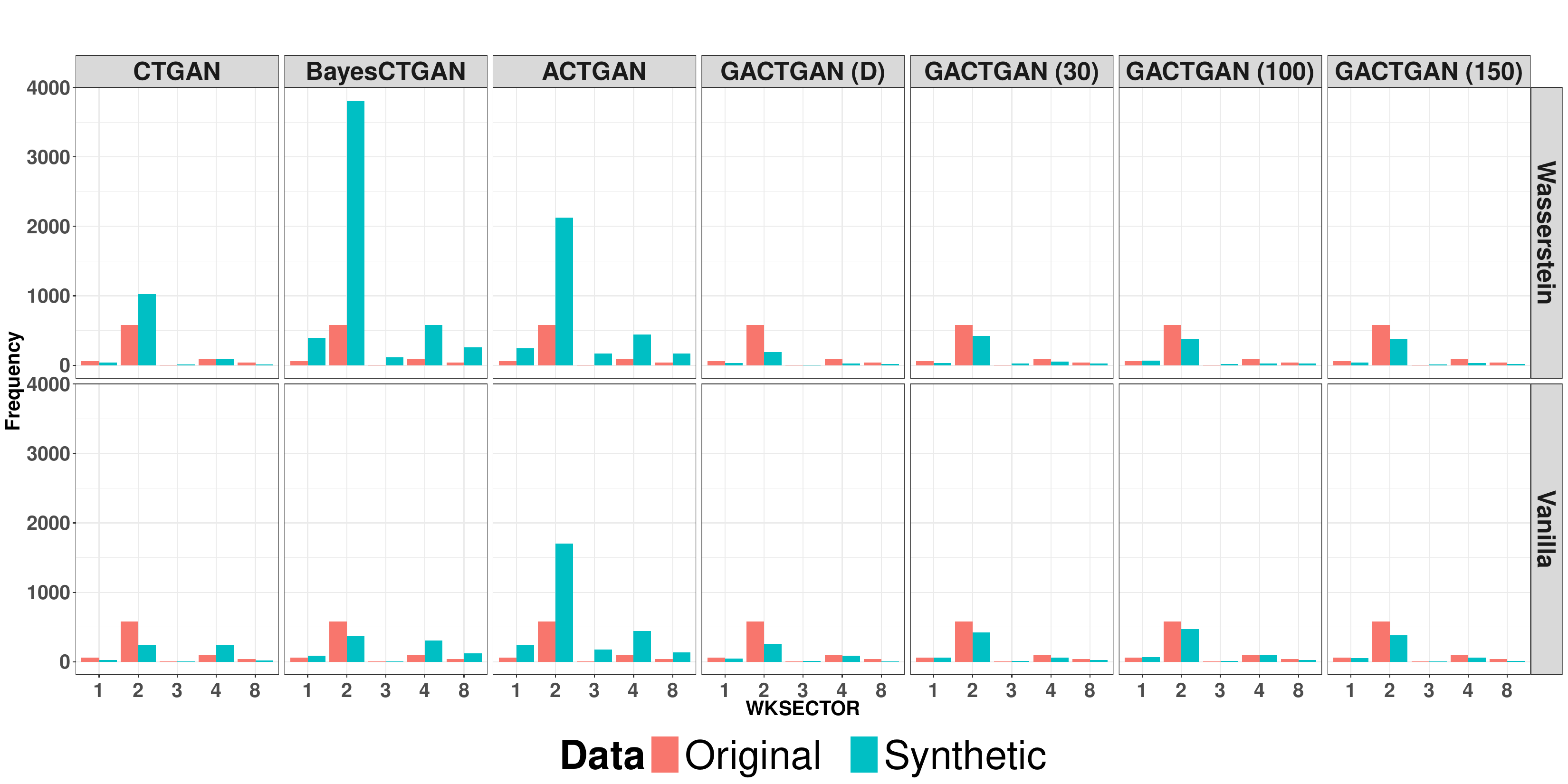}
    } \\ 
    \subfloat[CH-Customer exit among males]{
      \includegraphics[width=0.5\textwidth]{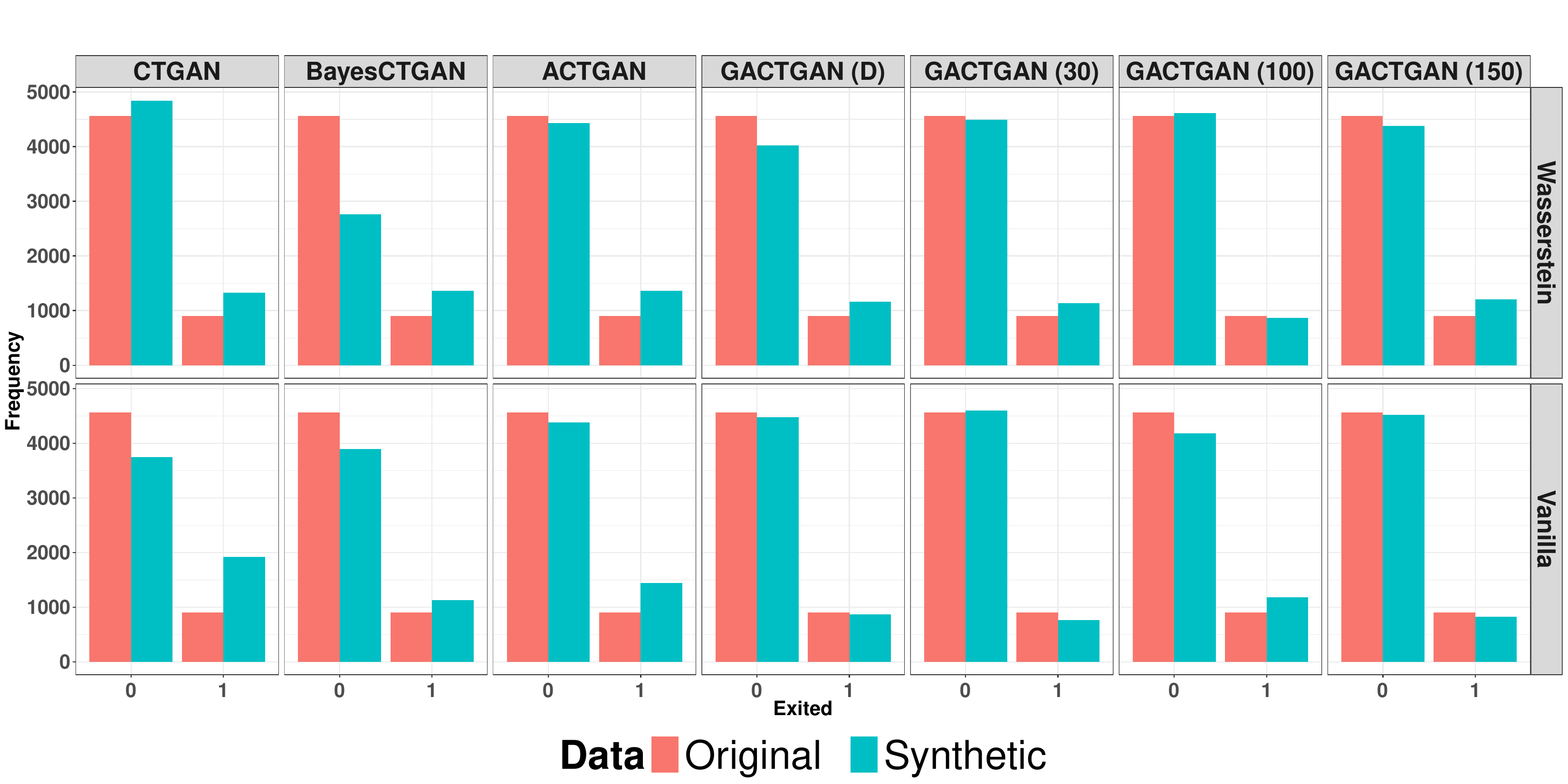}
    } 
    \subfloat[CH-Customer exit among females]{
      \includegraphics[width=0.5\textwidth]{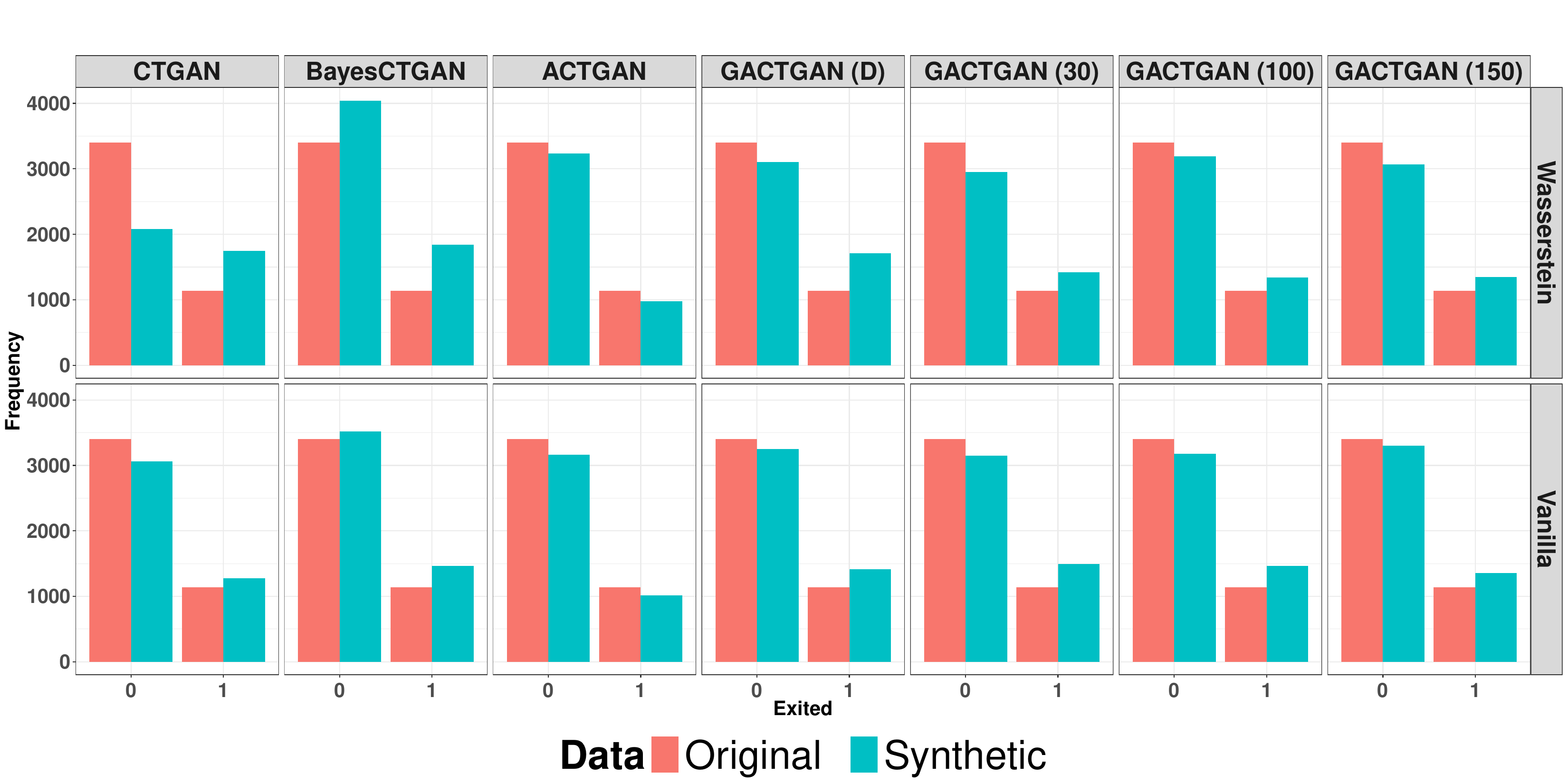}
    } \\
    \caption{Additional cross-tabulation between synthetic data and original datasets.}
    \label{fig:app-quali-ct}
\end{sidewaysfigure}

Figure~\ref{fig:app-quali-ct} reveals a further cross-tabulation in RW and CH. The analysis demonstrates GACTGAN's ability to accurately capture nuances in cross-tabulations, including employment sector patterns and gender-based customer exits. On the other hand, CTGAN and BayesCTGAN sometimes display inconsistent performance. These findings suggest that although GACTGAN is generally preferred for precise synthetic data generation, CTGAN could also yield good results even in smaller datasets.

\begin{sidewaysfigure}
    \centering
    \subfloat[AD-Age]{
      \includegraphics[width=0.5\textwidth]{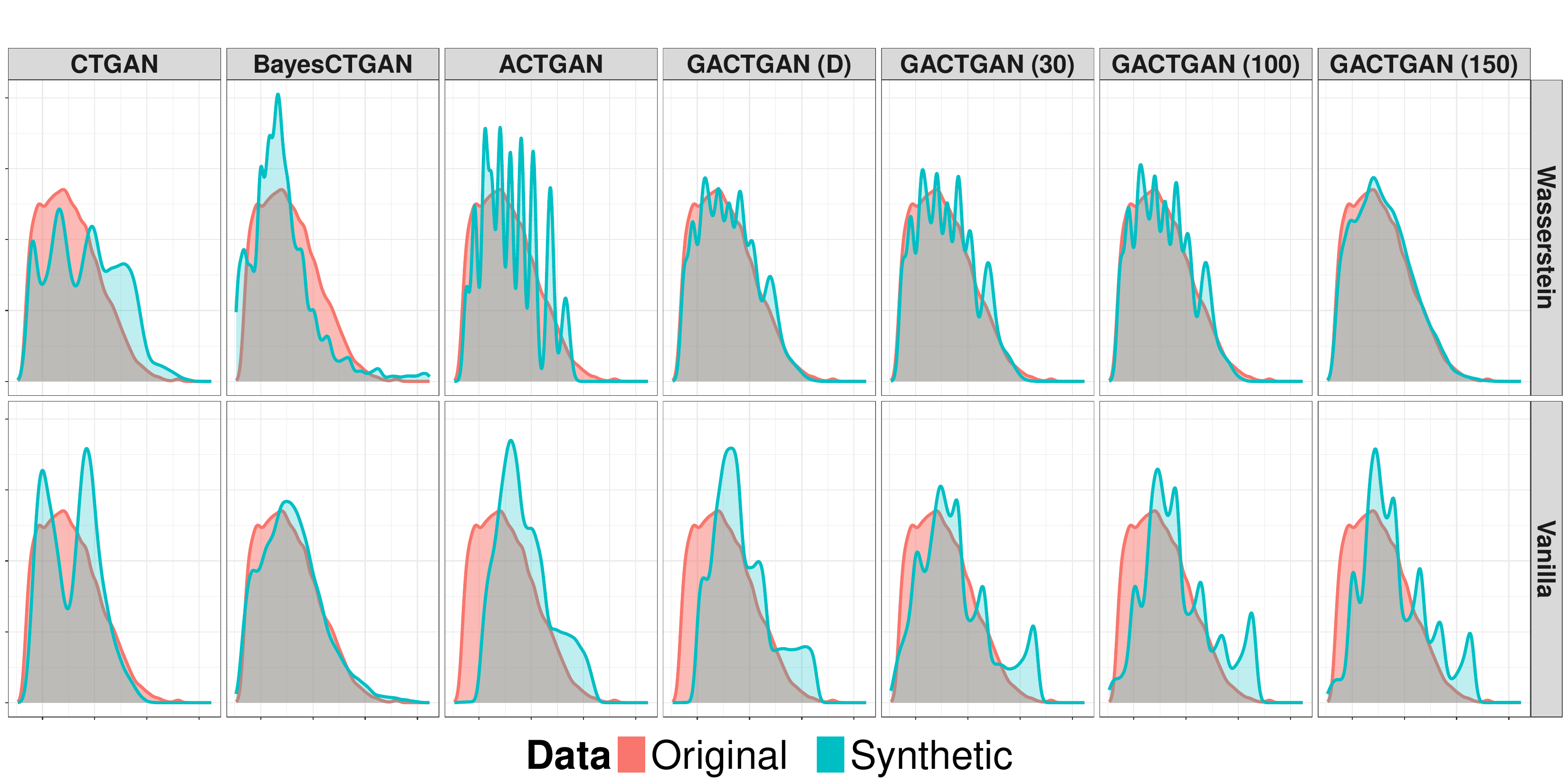}
    } 
    \subfloat[AD-fnlwgt]{
      \includegraphics[width=0.5\textwidth]{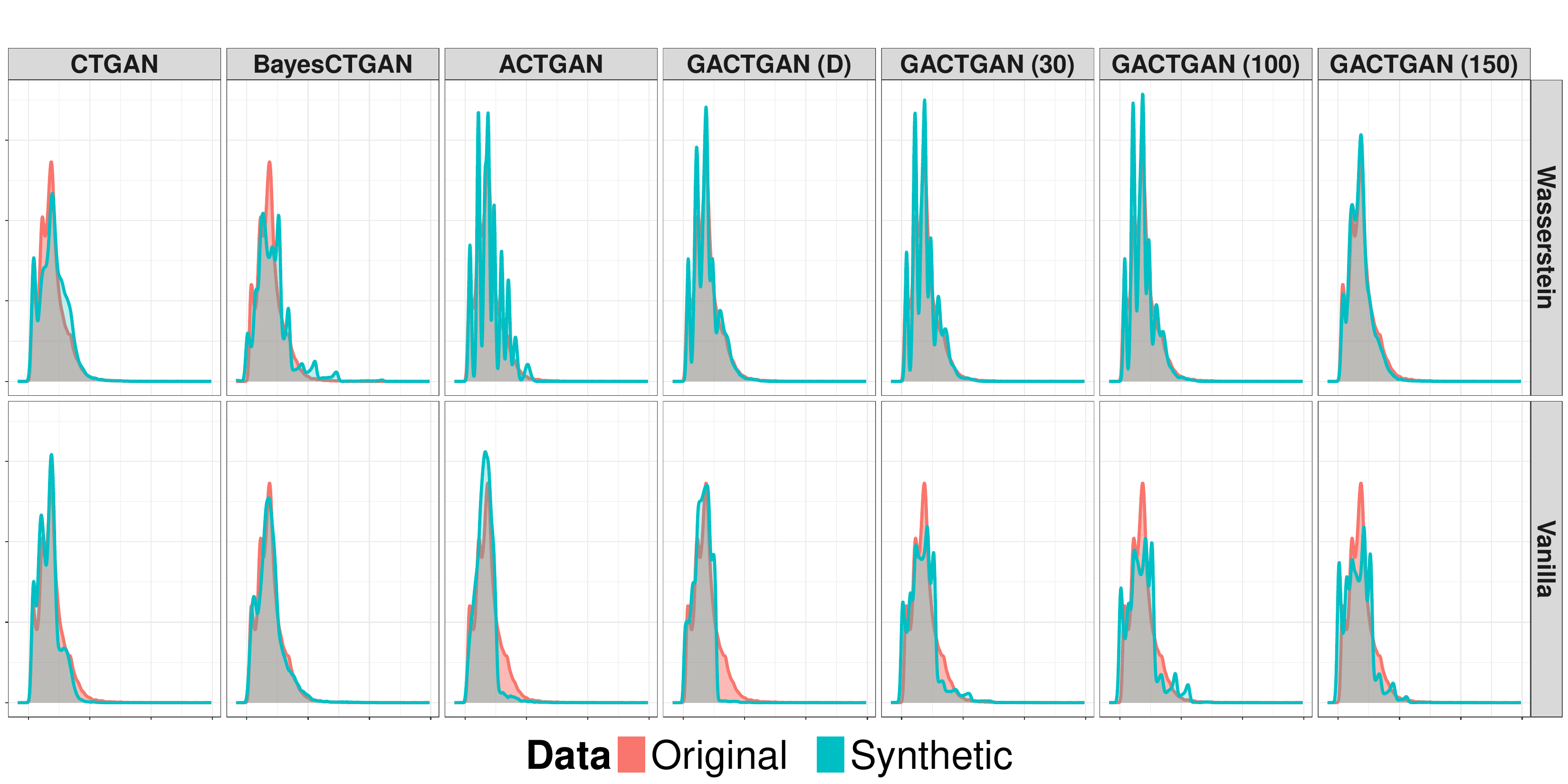}
    } \\ 
    \subfloat[CH-Age]{
      \includegraphics[width=0.5\textwidth]{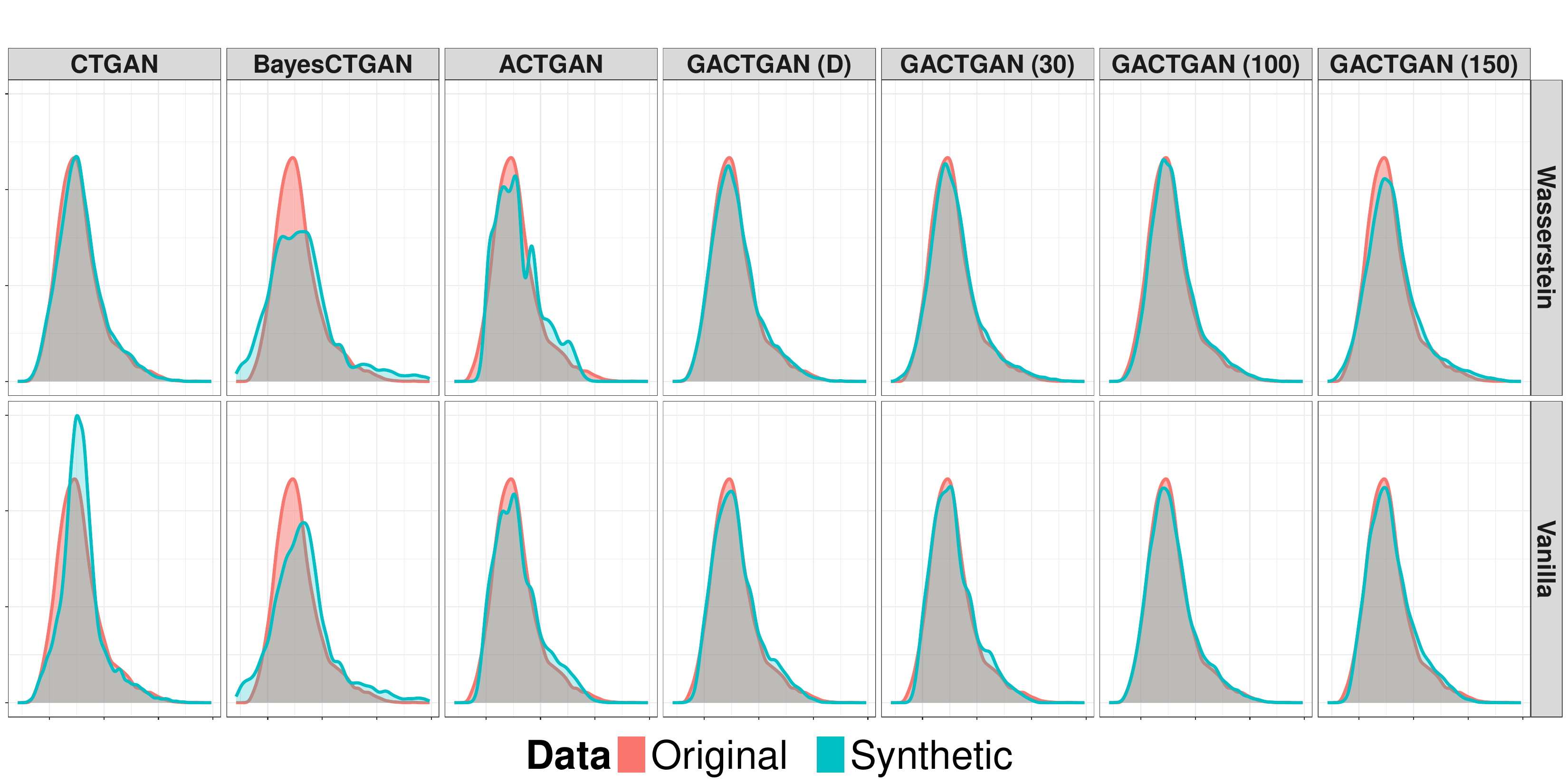}
    } 
    \subfloat[CH-Balance]{
      \includegraphics[width=0.5\textwidth]{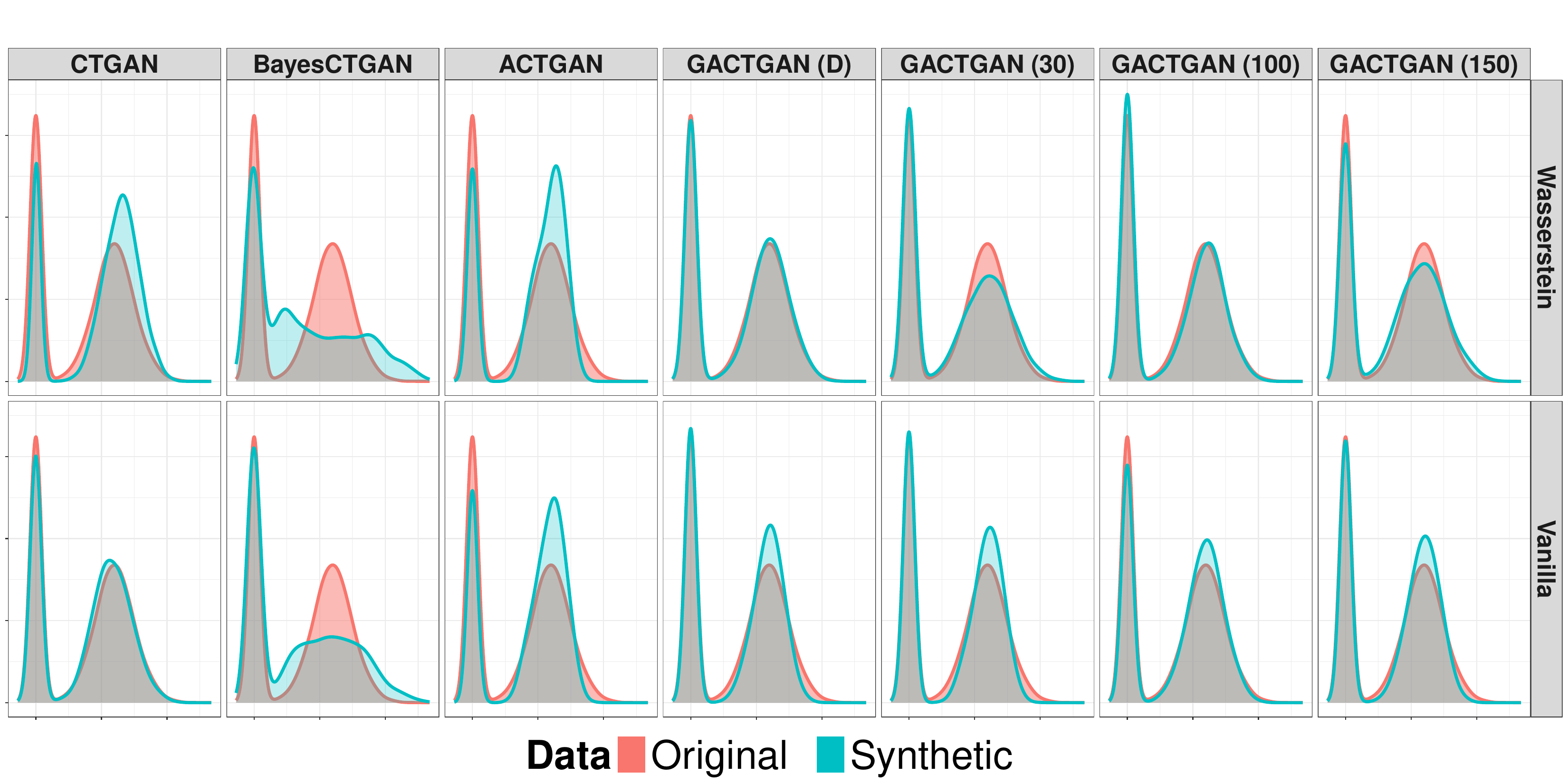}
    } \\
    \caption{Additional histogram between synthetic data and original datasets.}
    \label{fig:app-quali-hist}
\end{sidewaysfigure}

Figure~\ref{fig:app-quali-hist} presents additional histograms of continuous variables in AD and CH. The results suggest that GACTGAN frequently aligns more closely with the original data than other methods, indicating its potential to learn data distributions across different features. Nevertheless, in some cases, notably with variables like fnlwgt and Balance, CTGAN occasionally becomes competitive with GACTGAN in replicating the original distributions. These findings imply that GACTGAN generally has a greater capacity to reflect the complexities of datasets in certain contexts or with specific data types.

\begin{figure}
    \centering
    \subfloat[AD]{
      \includegraphics[width=\textwidth]{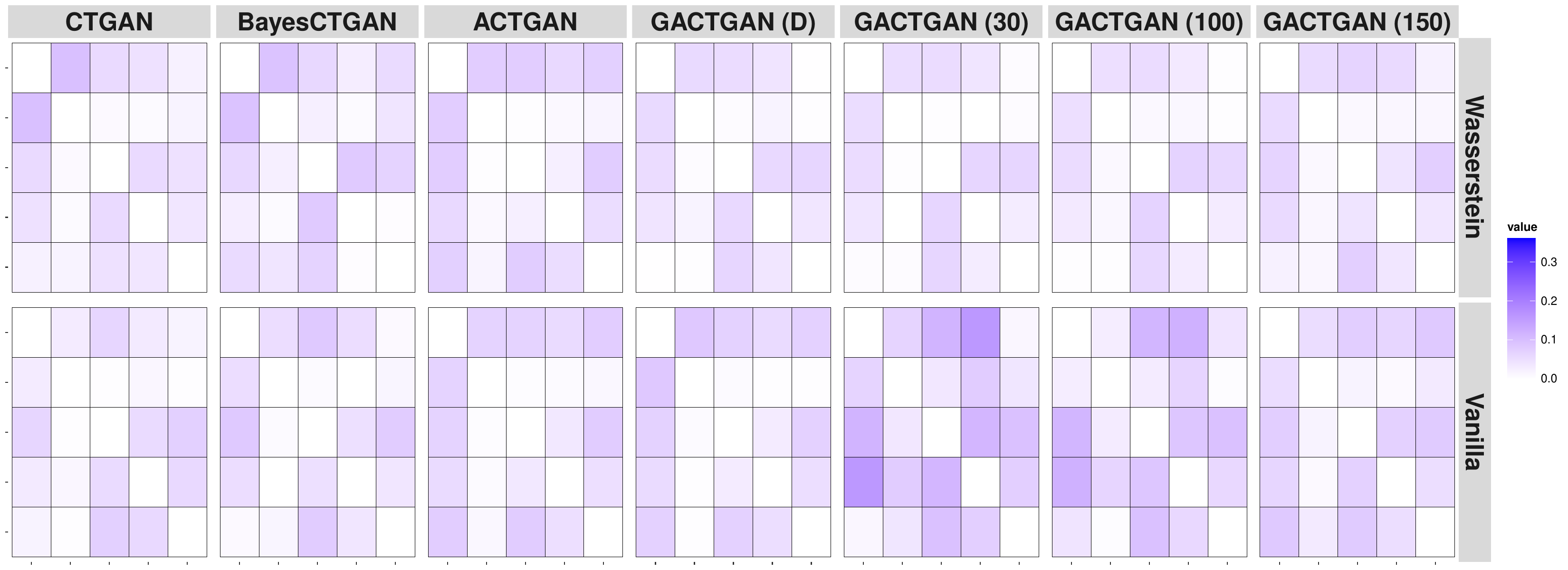}
    } \\
    \subfloat[CH]{
      \includegraphics[width=\textwidth]{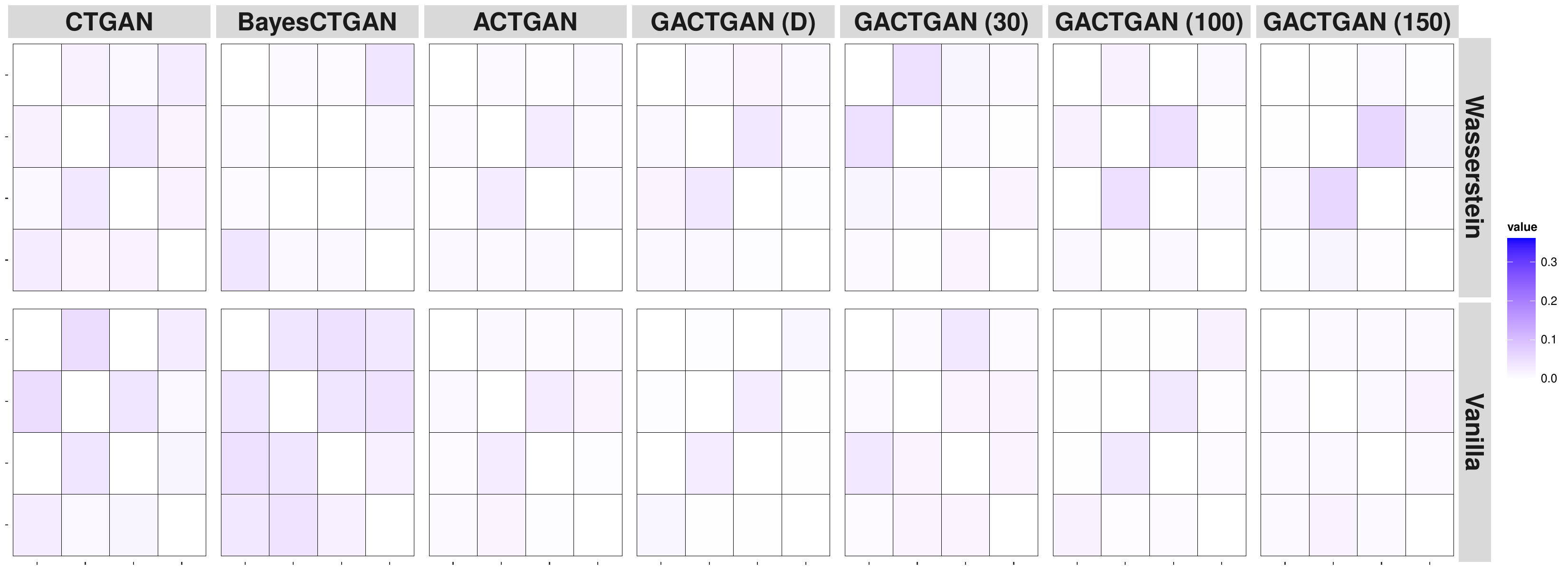}
    } 
    \caption{Additional correlation difference between synthetic and original dataset.}
    \label{fig:app-quali-corr}
\end{figure}

Figure~\ref{fig:app-quali-corr} presents additional correlation differences in AD and CH datasets. The results suggest that GACTGAN frequently could perform better by having a lower correlation difference in CH, whereas in AD the correlation difference is quite in a contest with CTGAN. These findings imply that GACTGAN generally has a good ability to preserve correlation.

\end{document}